\documentclass[pdflatex,sn-apa]{sn-jnl}

\usepackage{natbib}
\usepackage{amssymb}
\usepackage{color,soul}
\usepackage{aliascnt}
\usepackage{amsmath}
\usepackage{textgreek}
\usepackage{caption}
\usepackage{hyperref}
\usepackage{siunitx}
\usepackage{graphicx}
\usepackage{multirow}
\usepackage{adjustbox}
\usepackage{xcolor}
\usepackage{float}
\restylefloat{table}

\newaliascnt{eqfloat}{equation}
\newfloat{eqfloat}{h}{eqflts}
\floatname{eqfloat}{Equation}
\newcommand*{\ORGeqfloat}{}
\let\ORGeqfloat\eqfloat
\def\eqfloat{%
  \let\ORIGINALcaption\caption
  \def\caption{%
    \addtocounter{equation}{-1}%
    \ORIGINALcaption
  }%
  \ORGeqfloat
}

\jyear{2022}%

\theoremstyle{thmstyleone}%
\theoremstyle{thmstyletwo}%

\theoremstyle{thmstylethree}%

\begin{document}

\title[Active Learning and Novel Model Calibration Measurements in Manufacturing]{Active Learning and Novel Model Calibration Measurements for Automated Visual Inspection in Manufacturing}


\author*[1,2,5]{\fnm{Jo\v{z}e M.} \sur{Ro\v{z}anec}}\email{joze.rozanec@ijs.si}
\author[2]{\fnm{Luka} \sur{Bizjak}}\email{luka.bizjak@ijs.si}
\author[3]{\fnm{Elena} \sur{Trajkova}}\email{trajkova.elena.00@gmail.com}
\author[2]{\fnm{Patrik} \sur{Zajec}}\email{patrik.zajec@ijs.si}
\author[4]{\fnm{Jelle} \sur{Keizer}}\email{jelle.keizer@philips.com}
\author[5]{\fnm{Bla\v{z}} \sur{Fortuna}}\email{blaz.fortuna@qlector.com}
\author[2]{\fnm{Dunja} \sur{Mladeni\'{c}}}\email{dunja.mladenic@ijs.si}

\affil[1]{\orgname{Jo\v{z}ef Stefan International Postgraduate School}, \orgaddress{\street{Jamova 39}, \city{Ljubljana}, \postcode{1000}, \country{Slovenia}}}

\affil[2]{\orgname{Jo\v{z}ef Stefan Institute}, \orgaddress{\street{Jamova 39}, \city{Ljubljana}, \postcode{1000}, \country{Slovenia}}}

\affil[3]{\orgdiv{Faculty of Electrical Engineering}, \orgname{University of Ljubljana}, \orgaddress{\street{Tr\v{z}a\v{s}ka c. 25}, \city{Ljubljana}, \postcode{1000},  \country{Slovenia}}}

\affil[4]{\orgname{Philips Consumer Lifestyle BV}, \orgaddress{\street{Oliemolenstraat 5}, \city{Drachten}, \country{The Netherlands}}}

\affil[5]{\orgname{Qlector d.o.o.}, \orgaddress{\street{Rov\v{s}nikova 7}, \city{Ljubljana}, \postcode{1000}, \country{Slovenia}}}


\abstract{Quality control is a crucial activity performed by manufacturing enterprises to ensure that their products meet quality standards and avoid potential damage to the brand's reputation. The decreased cost of sensors and connectivity enabled increasing digitalization of manufacturing. In addition, artificial intelligence enables higher degrees of automation, reducing overall costs and time required for defect inspection. This research compares three active learning approaches, having single and multiple oracles, to visual inspection.
Six new metrics are proposed to assess the quality of calibration without the need for ground truth. Furthermore, this research explores whether existing calibrators can improve their performance by leveraging an approximate ground truth to enlarge the calibration set.
The experiments were performed on real-world data provided by \textit{Philips Consumer Lifestyle BV}. Our results show that the explored active learning settings can reduce the data labeling effort by between three and four percent without detriment to the overall quality goals, considering a threshold of p=0.95. Furthermore, the results show that the proposed calibration metrics successfully capture relevant information otherwise available to metrics used up to date only through ground truth data. Therefore, the proposed metrics can be used to estimate the quality of models' probability calibration without committing to a labeling effort to obtain ground truth data.}




\maketitle

\section{Introduction}\label{INTRODUCTION}

Quality control is one of the key parts of the manufacturing process, which comprehends inspection, testing, and identification to ensure the manufactured products comply with specific standards and specifications \citep{yang2020using,kurniati2015quality,wuest2014approach}. For example, the inspection tasks aim to determine whether a specific part features assembly integrity, surface finish, and adequate geometric dimensions \citep{newman1995survey}. In addition, product quality is key to the business since it (i) builds trust with the customers, (ii) boosts customer loyalty, and (iii) reinforces the brand reputation.

One such quality inspection activity is the visual inspection, considered a bottleneck activity in some instances \citep{zheng2020defect}. Visual inspection is associated with many challenges. Some visual inspections require a substantial amount of reasoning capability, visual abilities, and specialization \citep{newman1995survey}. Furthermore, reliance on humans to perform such tasks can affect the scalability and quality of the inspection. When considering scalability, human inspection requires training inspectors to develop inspection skills, their inspection execution tends to be slower when compared to machines, they fatigue over time and can become absent at work (due to sickness or other motives) \citep{vergara2014automatic,selvi2017effective}. The quality of inspection is usually affected by the inherent subjectiveness of each human inspector, the task complexity, the job design, the working environment, the inspectors' experience, well-being, and motivation, and the management's support and communication \citep{see2012visual,cullinane2013job,kujawinska2016role}. Manual visual inspection's scalability and quality shortcomings can be addressed through an automated visual inspection.

Automated visual inspection can be realized with Machine Learning models. Technological advances (e.g., Internet of Things or Artificial Intelligence \citep{rai2021machine,zheng2021applications}), and trends in manufacturing (e.g., the Industry 4.0 and Industry 5.0 paradigms \citep{rovzanec2022human}) have enabled the timely collection of data and foster the use of machine learning models to automate manufacturing tasks while reshaping the role of the worker \citep{carvajal2019online,chouchene2020artificial}. Automated visual inspection was applied in several use cases in the past \citep{duan2012machine,jiang2018fundamentals,villalba2019deep,beltran2020external}. Nevertheless, it is considered that the field is still in its early stages and that artificial intelligence has the potential to revolutionize product inspection \citep{aggour2019artificial}.

While machine learning models can be trained to determine whether a manufactured piece is defective and do so in an unsupervised or supervised manner, no model is perfect. At least three challenges must be faced: (a) how to improve the models' discriminative capabilities over time, (b) how to calibrate the models' prediction scores into probabilities to enable the use of standardized decision rules \citep{silva2021classifier}, and (c) how to alleviate the manual labeling effort. 

This paper presents our approach to addressing these three challenges as follows. Active learning is used to enhance the classification model to address the first challenge. Pool-based and stream-based settings are compared, considering different active learning sample query strategies across five machine learning algorithms. Platt scaling, a popular probability calibration technique, addresses the second challenge. Finally, two scenarios were considered when addressing the reduction of manual labeling effort: (i) manual inspection of cases where the machine learning model does not predict with enough confidence and (ii) data labeling to acquire ground truth data for the model calibration. The first scenario was addressed by exploring the usage of multiple oracles and soft labeling to reduce the manual inspection effort. Finally, the second scenario was addressed by approximating the ground truth with models' predictions to calibrate the model. Furthermore, several novel metrics to measure the quality of calibration were proposed. The results confirm that they can measure the quality of such calibration without needing a ground truth.

This work extends our previous research described in paper \textit{Streaming Machine Learning and Online Active Learning for Automated Visual Inspection} \citep{rovzanec2022streaming}. In that paper, research was performed to measure the impact of active learning on streaming algorithms. This paper explored batch and online settings along with different active learning policies and oracles. This research overcomes some of the shortcomings of the previous research.
First, it does not only consider the models' uncertainty to derive data instances to oracles but rather a certain quality acceptance level. Second, it calibrates the machine learning models so that through probability calibration, they issue probabilities rather than predictive scores. Third, it increases the amount of data devoted to active learning to ensure more meaningful results. Finally, it focuses on batch machine learning models (which achieve a greater discriminative performance) and studies them in batch and streaming active learning settings. In addition to the abovementioned items, multiple metrics were developed to assess the calibration quality of a calibrator. The metrics overcome some shortcomings of widely adopted metrics and enable measuring calibration quality when no ground truth is available. The research was performed on a real-world use case with images provided by \textit{Philips Consumer Lifestyle BV} corporation. The dataset comprises images regarding the printed logo on manufactured shavers. The images are classified into three classes: good prints, double prints, and interrupted prints.

The Area Under the Receiver Operating Characteristic Curve (AUC ROC, see \citep{BRADLEY19971145}) was used to evaluate the discriminative capability of the classification models. AUC ROC estimates the quality of the model for all possible cutting thresholds. It is invariant to \textit{a priori} class probabilities and, therefore, suitable for classification tasks with strong class imbalance. Furthermore, given that the models were evaluated in a multiclass setting, the AUC ROC was computed with a one-vs-rest strategy. Furthermore, the performance of multiple probability calibration approaches was measured through the Estimated Calibration Error (ECE) and several novel metrics proposed in this research. 

This paper is organized as follows. Section \ref{S:RELATED-WORK} describes the current state of the art and related works. Section \ref{S:PROBABILITIES-CALIBRATION} describes novel metrics proposed for probability calibration and how calibration methods can leverage approximate ground truth to enlarge the calibration set. The main novelty regarding the proposed probabilities calibration metrics is the ability to measure calibration quality without needing a ground truth. Section \ref{S:USE-CASE} describes the use case, while Section \ref{S:METHODOLOGY} provides a detailed description of the methodology  followed. Section \ref{S:EXPERIMENTS} describes the experiments performed, while Section \ref{S:RESULTS-AND-EVALUATION} presents and discusses the results obtained. Finally, Section \ref{S:CONCLUSION} presents the conclusions and outlines future work.

\section{Related Work}\label{S:RELATED-WORK}
This section provides a short overview of three topics relevant to this research: (i) the use of machine learning for quality inspection, (ii) active learning, and (iii) probabilities calibration. The following  subsections are devoted to each of them.

\subsection{Machine Learning for Quality Inspection}

A comprehensive and reliable quality inspection is often indispensable to the manufacturing process, and high inspection volumes turn inspection processes into bottlenecks \citep{schmitt2020predictive}. Machine Learning has been recognized as a technology that can drive the automation of quality inspection tasks in the industry. Multiple authors report applying it for early prediction of manufacturing outcomes, which can help drop a product that will not meet quality expectations and avoid investment in expensive manufacturing stages. Furthermore, similar predictions can be used to determine whether the product can be repaired and therefore avoid either throwing away a piece to which the manufacturing process was invested or selling a defective piece with the corresponding costs for the company \citep{weiss2016continuous}. Automated visual inspection refers to image processing techniques for quality control, usually applied in the production line of manufacturing industries \citep{beltran2020external}. It has been successfully applied to determine the end quality of the products. It provides many advantages, such as performing non-contact inspection that is not affected by the type of target, surface, or ambient conditions (e.g., temperature) \citep{park2016machine}. In addition, visual inspection systems can perform multiple tasks simultaneously, such as object, texture, or shape classification and defect segmentation, among other inspections. Nevertheless, automated visual inspection is a challenging task given that collecting the dataset is usually expensive, and the methods developed for that purpose are dataset-dependent \citep{ren2017generic}.

\citep{jian2017automatic} considers three approaches that exist toward automated visual inspection: (a) classification, (b) background reconstruction and removal (reconstruct and remove background to find defects in the residual image), and (c) template reference (comparing a template image with a test image). \citep{tsai2008defect} describe how TFT-LCD panels and LCD color filters were inspected by comparing surface segments containing complex periodic patterns. \citep{kang2005surface} successfully applied feed-forward networks to detect surface defects on cold-rolled strips. In the same line, \citep{yun2014defect} proposed a novel defect detection algorithm for steel wire rods produced by the hot rolling process. \citep{valavanis2010multiclass} compared multiple machine learning models (Support Vector Machine, Neural Network, and K-nearest neighbors (kNN)) on defect detection in weld images. \citep{park2016machine} developed a Convolutional Neural Network (CNN) and compared it to multiple models (particle swarm optimization-imperialist competitive algorithm, Gabor-filter, and random forest with variance-of-variance features) to find defects on silicon wafers, solid paint, pearl paint, fabric, stone, and wood surfaces. Multiple authors developed machine learning algorithms for visual inspection leveraging feature extraction from pre-trained models \citep{cohen2020sub,li2021cutpaste,jezek2021deep}. While much research was devoted to supervised machine learning methods, unsupervised defect detection was explored by many authors, who explored using Fourier transforms to remove regularities and highlight irregularities (defects) \citep{aiger2012phase} or employed autoencoders to find how a reference image differs from the expected pattern \citep{mujeeb2018unsupervised,zavrtanik2021draem}.

\subsection{Active Learning}
Active learning is a subfield of machine learning that studies how an active learner can best identify informative unlabeled instances and requests their labels from some \textit{oracle}. Typical scenarios involve (i) membership query synthesis (a synthetic data instance is generated), (ii) stream-based selective sampling (the unlabeled instances are drawn one at a time, and a decision is made whether a label is requested or the sample is discarded), and (iii) pool-based selective sampling (queries samples from a pool of unlabeled data). Among the frequently used querying strategies are (i) uncertainty sampling (select an unlabeled sample with the highest uncertainty, given a certain metric, or machine-learning model \citep{lewis1994heterogeneous}), or (ii) query-by-committee (retrieve the unlabeled sample with the highest disagreement between a set of forecasting models (\textit{committee}) \citep{cohn1994improving,settles2009active}) can be found. More recently, new scenarios have been proposed leveraging reinforcement learning, where an agent learns to select images based on their similarity, and rewards obtained are based on the oracle's feedback \citep{ren2020survey}. In addition, it has been demonstrated that ensemble-based active learning can effectively counteract class imbalance through newly labeled image acquisition \citep{beluch2018power}. While active learning reduces the required volume of labeled images, it is also essential to consider that it can produce an incomplete ground truth by missing the annotations of defective parts classified as false negatives and not queried by the active learning strategy \citep{cordier2021active}.

Active learning was successfully applied in manufacturing, but scientific literature remains scarce on this domain \citep{meng2020machine}. Some use cases include the automatic optical inspection of printed circuit boards \citep{dai2018towards}, media news recommendation in a demand forecasting setting \citep{zajec2021towards}, and the identification of the local displacement between two layers on a chip in the semiconductor industry \citep{van2018active}.

\subsection{Probabilities calibration}
Probabilities are defined as a generalization of predicate calculus where given a truth value of a formula given the evidence (degree of plausibility) is generalized to a real number between zero and one \citep{cheeseman1985defense}. Many machine learning models output prediction scores which cannot be directly interpreted as probabilities. Therefore, such models can be calibrated (mapped to a known scale with known properties), ensuring the prediction scores are converted to probabilities. Probability calibration aims to provide reliable estimates of the true probability that a sample is a member of a class of interest. Such calibration (a) usually does not decrease the classification accuracy, (b) enables using provides thresholds on the decision rules and therefore minimizes the classification error, (c) ensures decision rules and their maximum posterior probability are fully justified from the theoretical point of view, (d) can be easily adapted to changes in class and cost distributions, and therefore (e) is key to decision-making tasks \citep{cohen2004properties,song2021classifier}.

The k-class probabilistic classifier is considered well-calibrated if the predicted k-dimensional probability vector has a distribution that approximates the distribution of the test instances. A different criterion is introduced with confidence calibration, aiming only to calibrate the classifier's most likely predicted class \citep{song2021classifier}. 

Multiple probability calibration methods have been proposed in the scientific literature. The post-hoc techniques aim to learn a calibration map for a machine-learning model based on hold-out validation data. In addition, popular calibration methods for binary classifiers include logistic calibration (Platt scaling), isotonic calibration, Beta calibration, temperature calibration, and binning calibration. 

Empirical binning builds the calibration map by computing the empirical frequencies within a set of score intervals. It can therefore capture arbitrary prediction score distributions \citep{kumar2019verified}. 
Isotonic regression computes a regression assuming the uncalibrated model has a set of non-decreasing constant segments corresponding to bins of varying widths. Given its non-parametric nature, it avoids a model misfit, and due to the monotonicity assumption, it can find optimal bin edges. Nevertheless, training times and memory consumption can be high on large datasets and give sub-optimal results if the monotonicity assumption is violated.
Platt scaling \citep{platt20005} aims to transform prediction scores into probabilities through a logistic regression model, considering a uniform probability vector as the target. While the implementation is straightforward and the training process is fast, it assumes the input values correspond to a real scalar space and restricts the calibration map to a sigmoid shape. 
Probability calibration trees evolve the concept of Platt scaling, identifying regions of the input space that lead to poor probability calibration and learning different probability calibration models for those regions, achieving better overall performance \citep{leathart2017probability}.
Beta calibration was designed for probabilistic classifiers. It assumes that the scores of each class can be approximated with two Beta distributions and is implemented as a bivariate logistic regression.
Temperature scaling uses a scalar parameter $T>0$ (where T is considered the temperature) to rescale logit scores before applying a softmax function to achieve recalibrated probabilities with better spread scores between zero and one. It is frequently applied to deep learning models, where the prediction scores are frequently strongly skewed towards one or zero. Furthermore, the method can be applied to generic probabilistic models by transforming the prediction scores with a logit transform \citep{guo2017calibration}. This enables calculating the score against a reference class and obtaining the ratio against other classes. Nevertheless, the method is not robust in capturing epistemic uncertainty \citep{ovadia2019can}. Finally, the concept of temperature scaling is extended in vector scaling, which considers that a different temperature for each class can be specified, and matrix scaling, which considers a matrix and intercept parameters \citep{song2021classifier}.

Several metrics and methods were proposed to assess the quality of the calibration. Reliability diagrams plot the observed relative frequency of predicted scores against their values. They, therefore, enable to quickly assess whether the event happens with a relative frequency consistent with the forecasted value \citep{brocker2007increasing}. On the other hand, validity plots aim to convey the bin frequencies for every bin and therefore provide valuable information regarding miscalibration bounds \citep{gupta2021distribution}. Among the metrics, the binary ECE measures the average gap across all bins in a reliability diagram, weighted by the number of instances in each bin, considering the labeled samples of a test set. In the same line, the binary Maximum Calibration Error computes the maximum gap across all bins in a reliability diagram. The Confidence Estimated Calibration Error measures the average difference between accuracy and average confidence across all bins in a confidence reliability diagram, weighted by the number of instances per bin. A different approach is followed by the Brier score, which measures the mean squared difference between the predicted probability and the actual outcome. While the ECE metric is widely accepted, research has shown is subject to shortcomings \citep{nixon2019measuring,posocco2021estimating}. One of such shortcomings is that when using fixed calibration ranges, some bins contain most of the data, resulting in the metric's decreased sharpness. Furthermore, ECE is measured across non-empty bins, failing to account for the overall distribution of positives across the mean predicted probabilities. Measuring probabilistic calibration remains a challenge \citep{nixon2019measuring}.

While many probability calibration methods and metrics have been developed, most of them were conceived considering probability calibration must be done based on some ground truth. Nevertheless, acquiring data for such ground truth is expensive (requires labeled instances), limits the amount of data seen to build such a probability calibration map, and therefore introduces inaccuracies due to the inherent characteristics of the sample. To address this void, this research proposes labeling each predicted data instance according to the predicted class with the highest score or most likely class if the highest predicted scores are equal. Assuming the classifier could perform with perfect discriminative power in the best case, such labels would equal the ground truth. Furthermore, this research proposes metrics to assess the discrepancy between an ideal probability calibration scenario and the calibrated classifier to measure the quality of probability calibration achieved. By doing so, the calibrators' quality over time can be measured without needing any data labeling for such an assessment. Furthermore, it enables exploring approximate model's probabilities calibration, training a calibrator from a ground truth approximated with predicted labels. This idea is further explained and developed in Section \ref{S:PROBABILITIES-CALIBRATION}.

\section{Approximate Model's Probabilities Calibration}\label{S:PROBABILITIES-CALIBRATION}

\subsection{Towards Approximate Probability Calibration Models}

This research proposes metrics and an approach to calibrating machine learning prediction scores to probabilities by using a ground truth approximation. The approach considers building an initial calibration set, as it is common practice for probability calibration methods. A calibration set has (a) several prediction scores used to perform the probability calibration and (b) the ground truth labels for the corresponding data instances. Using both, a mapping is created between the prediction scores and the probability of a class outcome. Nevertheless, the limited amount of data in the calibration set can impact the fidelity of the calibration. In particular, the distribution of predictive scores between the calibration set and the predictions performed in a production environment can differ. 

The final prediction of a calibrated model has at least two sources of error: (a) the classification model, which does not perfectly predict the target class, and (b) the probability calibration technique, which does not produce a perfect probabilistic mapping between the predicted scores and the target class. While metrics and plots exist to assess the quality of the probability calibration, such means require a ground truth to evaluate the probability calibration. While the requirement for a ground truth allows for an exact estimate of the classifier on that particular hold-out data, it has at least two drawbacks: (i) it requires labeling a certain amount of data to perform the evaluation, and (ii) such data may not be representative of current or future data distributions observed in a production environment.

\begin{figure*}[!ht]
\centering
\includegraphics[width=\textwidth]{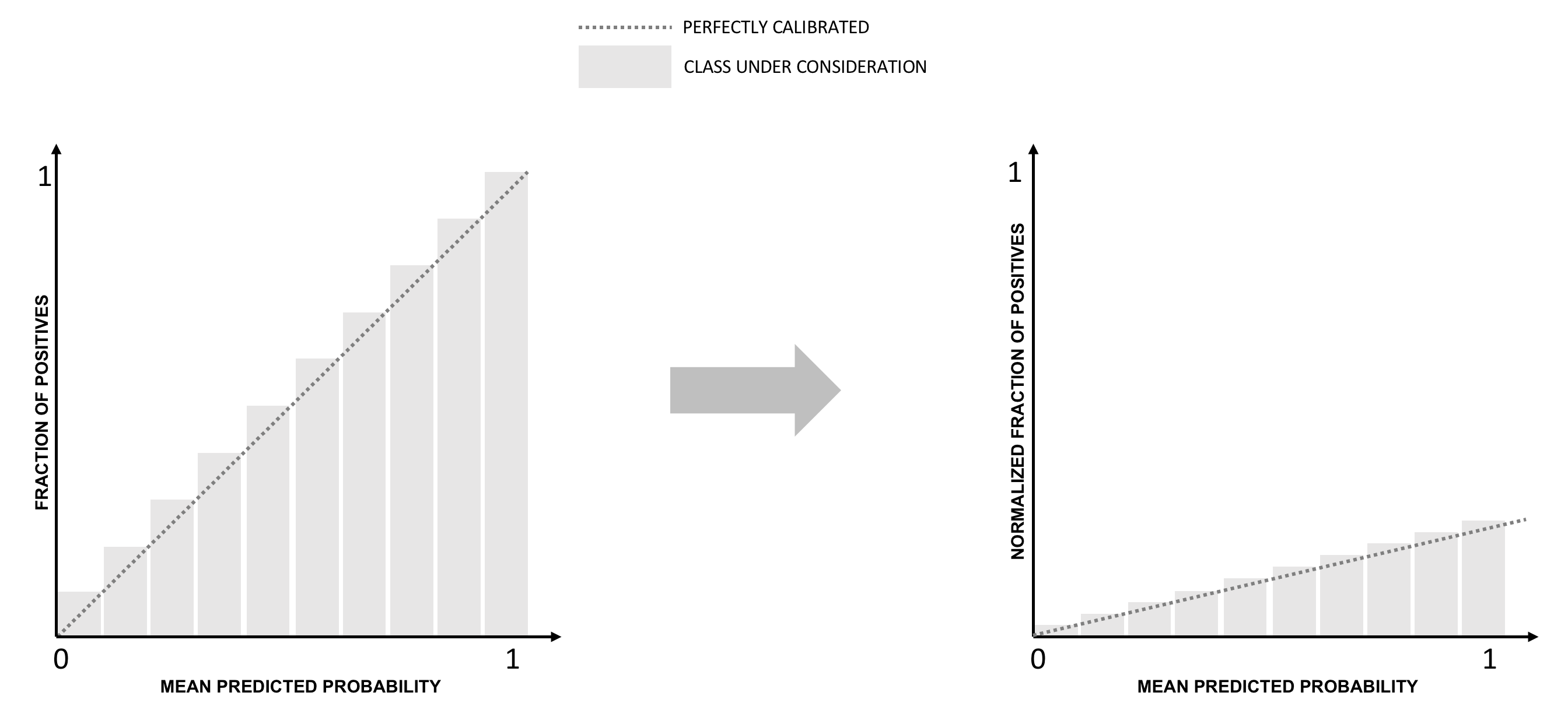}
\caption{The figure presents two calibration plots. On the left, the calibration plot shows a perfectly calibrated calibrator (where the fraction of positives for the class under consideration equals the mean predicted probability). On the right, the same information is presented, but normalizing the values of the plot on the right to ensure the sum of their values equals one.}
\label{F:CALIB-HISTOGRAM-CALIBRATION-DIAGONAL}
\end{figure*}

\subsection{Intuitions Behind a Calibration Without a Ground Truth}
Current scientific literature considers the quality of a model calibration can be measured by comparing, given a fixed class, whether the fraction of positives does correspond to the predicted mean probability of a given classifier. The fraction of positives empirically measures the likelihood of positive class events for the class under consideration within a specific mean probability range (bin). In a well-calibrated model, the likelihood of the occurrence of positive class events in a particular bin for the class under consideration matches the mean predicted probability, revealing a linear relationship between the mean predicted probability and the likelihood of the occurrence in that bin of the positive class event for the class considered (see Fig. \ref{F:CALIB-HISTOGRAM-CALIBRATION-DIAGONAL}). Furthermore, a perfectly calibrated classifier is only possible for a binary classification problem with no class imbalance. Class imbalance or multiple classes introduce distortions regarding the frequency with which the positive class is observed within a given predicted mean probability range compared to the frequency with the other events occurring within that mean probability range. 

For a well-calibrated classifier, each of the predicted classes is expected to behave as shown in Fig. \ref{F:CALIB-HISTOGRAM-CALIBRATION-DIAGONAL}. Therefore, while class imbalance or a multi-class setting can introduce distortions to the histogram's shape, the distance to the ideal case could be measured by comparing the histogram shape of a perfectly calibrated model for a given class and the shape of the histogram in the real-world case under consideration. To estimate how close the histograms are from each other, optimal transport is used \citep{villani2009optimal,peyre2019computational}. In particular, the Wasserstein distance measures the distance between the two histogram distributions. Nevertheless, the fraction of positives for a given class cannot be computed when no ground truth is available. Therefore, we reframe the problem so that the goodness of a model calibration can be evaluated even without considering a ground truth.

Considering the information available in Fig. \ref{F:CALIB-HISTOGRAM-CALIBRATION-DIAGONAL} and a particular class $j$, and considering each prediction regarding class $j$ an event $x$, we are interested on two types of events: $E_{1}=\{x \, corresponds \, to \, bin \, i \}$, and $E_{2}=\{x \, corresponds \, to \, class \, j \}$. Furthermore, we are interested in calibrating the model so that the resulting score indicates $p_{j}(E_{2} \rvert E_{1})$.

\subsubsection{Intuition 1: Considering a perfectly calibrated classifier}\label{SS:INTUITION-PERFECT-CALIBRATION}
Let us consider the case of a perfectly calibrated classifier. Given a perfectly calibrated classifier, the fraction of positives for a given class must match the mean predicted probability. The fraction of positives within a certain bin $i$ can be considered the empirical computation of $p_{j}(E_{2} \rvert E_{1})$. $E_{1}$ and $E_{2}$ are not independent events, given the probability of belonging to class $j$ should be higher in bins representing a higher mean predicted probability. Therefore, $p_{j}(E_{2} \rvert E_{1}) = \frac{p_{j}(E_{2} \cap E_{1})}{p_{j}(E_{1})}$. Considering a balanced binary classification problem, the number of predictions issued for each mean predicted probability range must be equal to verify the symmetry regarding the fraction of positives observed in the mean probability ranges for both classes. Fluctuations regarding the fraction of positives observed in the mean predicted probability ranges translate into an unequal number of predictions in them and directly impact the quality of the calibration. Based on this observation, given the abovementioned equation, $p_{j}(E_{2} \rvert E_{1}) = \frac{p_{j}(E_{2} \cap E_{1})}{p_{j}(E_{1})}$, $p_{j}(E_{1})$ is constant, and can be empirically computed as $p_{j}(E_{1}) = \frac{1}{\# \, of \, bins}$. The number of predictions for a given class $j$ is computed as the count of predictions where the highest predicted value was issued for that class $j$. While $p_{j}(E_{2} \cap E_{1})$ cannot be computed without ground truth, the expected values that must be satisfied for each bin for $p_{j}(E_{2} \rvert E_{1})$ are known. Therefore, we envision at least two ways to estimate the mismatch between the ideal case and the case under consideration. First, the value of $p_{j}(E_{2} \cap E_{1})$ can be inferred based on the expected $p_{j}(E_{2} \rvert E_{1})$ for a particular bin and the empirical computation of $p_{j}(E_{1})$ to then measure the Wasserstein distance between the resulting distributions. Second, it could be estimated by only considering $p_{j}(E_{1})$ and measuring the Wasserstein distance between the ideal distribution (an equal number of predictions per mean predicted probability range) and the distribution of predictions obtained from the calibrated classifier under consideration (number of predictions per bin, that are empirically measured - usually the amount of predictions is not equal across bins given the calibrated classifier's imperfection)). Each class's calibration quality could be estimated in both cases by comparing two histograms: the ideal case and the calibration model under consideration. The distance between both distributions computes measures how far the particular calibrator is from a perfectly calibrated case. 

While the case above was demonstrated for a balanced binary classification problem, it approximately holds for multiclass settings and cases with class imbalance. In these scenarios, we aim to calibrate each class as perfectly as possible, even though a perfect calibration cannot be achieved. Nevertheless, how well-calibrated each class is against the ideal case can still be assessed by comparing the distributions described above.

\subsubsection{Intuition 2: Considering a perfect classifier}\label{SS:INTUITION-PERFECT-CLASSIFIER}
Let us consider the case of a perfect classifier. Given a perfect classifier, the prediction equals the ground truth regarding a positive class event for the class under consideration. Therefore, two scenarios are considered: (a) degrade the classifiers' performance to achieve a calibrated classifier, or (b) spread the predicted values within a specific range so that they emulate particular calibration. It must be noted that while (a) can still satisfy the definition of probability considered for calibration, (b) does not.

For (a), the classifier's performance must be degraded due to the inherent definition of probabilities used in this problem: the calibration model will ensure a proportion of positive events regarding a class given a mean predicted probability bin. Therefore, given $n=number \, of \, classes$, the highest predicted value for each class will not issue only data instances of that class above $1/n$. Furthermore, some cases will be lost under the $1/n$ threshold.

\begin{figure*}[!ht]
\centering
\includegraphics[width=\textwidth]{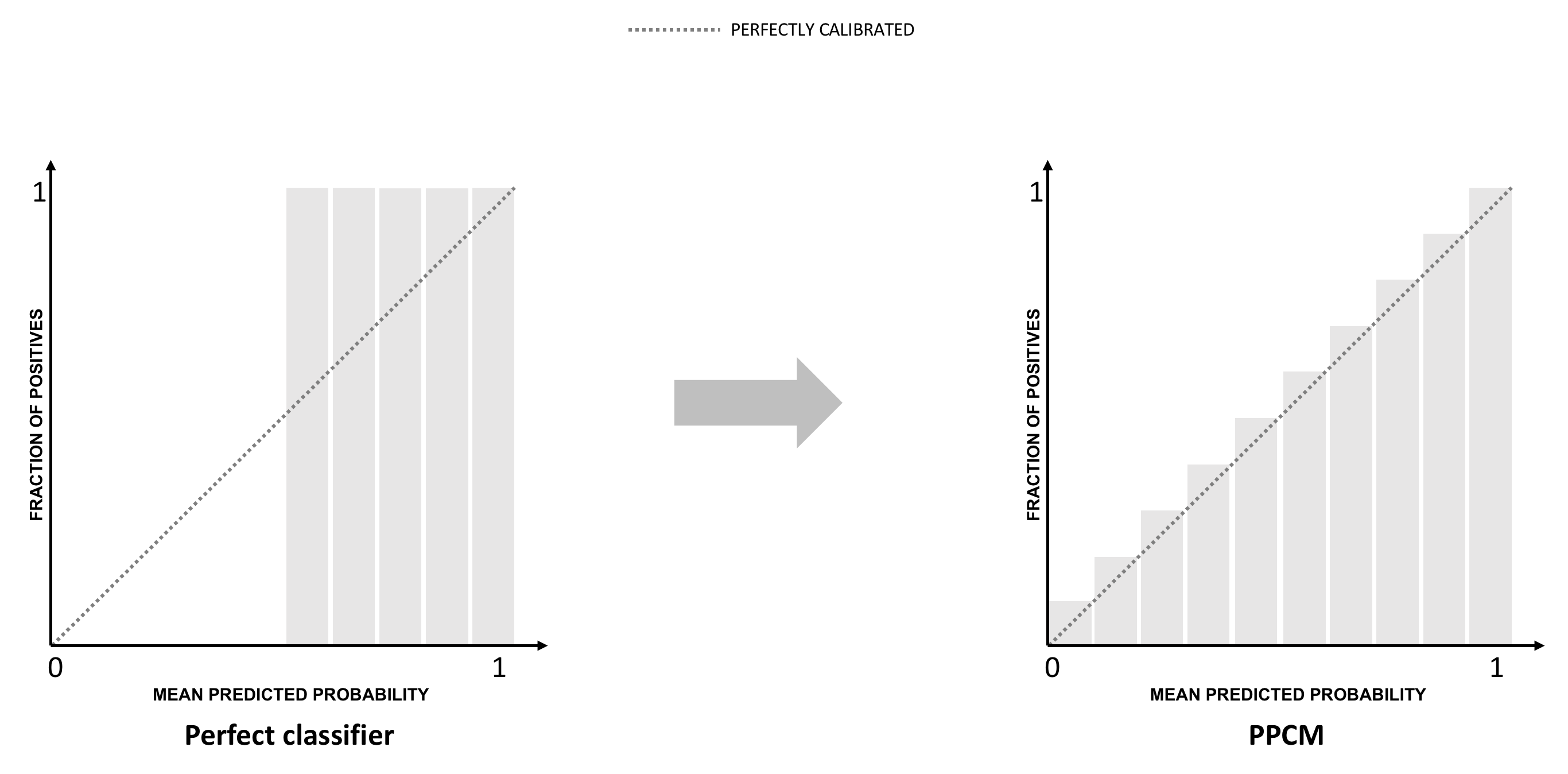}
\caption{The figure presents two calibration plots. On the left, the calibration plot shows a perfect binary classifier, while on the right we find a perfectly calibrated binary classifier.}
\label{F:CALIB-PERFECT-TO-CALIBRATED}
\end{figure*}

On the other hand, for (b), the abovementioned equation $p_{j}(E_{2} \rvert E_{1}) = \frac{p_{j}(E_{2} \cap E_{1})}{p_{j}(E_{1})}$ can be considered. It is known that for a perfect classifier, the following is true: $p_{j}(E_{2})=0$ or $p_{j}(E_{2})=1$. Furthermore, $E_{1}$ and $E_{2}$ can be considered dependent events, given $p_{j}(E_{2})=0$ for bins below a certain threshold, and $p_{j}(E_{2})=1$ otherwise (see Fig. \ref{F:CALIB-PERFECT-TO-CALIBRATED}). In addition, the mean predicted probability would not match the fraction of positives, given the classifier is perfect: each prediction perfectly identifies the target class. Therefore, this scenario \textit{per se} violates the idea behind probabilies calibration.  Nevertheless, the best approximation towards Fig. \ref{F:CALIB-HISTOGRAM-CALIBRATION-DIAGONAL} would be to achieve an increasing number of predictions per mean predicted probability range (histogram bin) for a specific class. To avoid degrading the models' discriminative power, such a mapping function will not issue scores below $1/n$, where $n=number \, of \, classes$. 

\subsection{Intuitions Materialized}
\subsubsection{From Intuitions to Approximate Calibrators}
To perform model calibration, a function that can map the predictive scores of a machine-learning model to probability scores is required. Ideally, such probability scores would indicate $p_{j}(E_{2} \rvert E_{1})$. When no ground truth is available, the intuitions described above can be considered to reproduce some scenarios where the resulting probability score distribution can be compared against an ideal probability score distribution. Therefore, we consider labeling the predicted data instances with the class with the highest predicted score. In case two classes hold equal scores, we decide on the most probable one based on the class imbalance observed in the train test. For balanced datasets, the class can be assigned randomly, given no other information exists to guide the decision. The more perfect the classification model, the closer will the assigned labels be to the ground truth. Given data instances with predicted scores and assigned labels, a calibrator can be fitted to map the classifier's output to a calibrated probability.

\subsubsection{From Intuitions to Metrics}
In subsections \ref{SS:INTUITION-PERFECT-CALIBRATION} and \ref{SS:INTUITION-PERFECT-CLASSIFIER}, the cases of a perfectly calibrated model and a perfect classifier were considered. While in the case of a perfect classifier, a ground truth is not needed (the predicted labels equal the ground truth), non-perfect classifiers approximate such a ground truth to a certain degree (measured as the classifiers' performance). Furthermore, regardless of the calibration technique, it was shown that a certain correlation between the calibration quality and the calibration score distribution exists. In particular, it was shown that for each class $k$ a histogram could be computed showing (a) the number of predictions per bin and (b) the proportion of positive class occurrences per mean predicted probability bin. Both could then be compared against ideal cases. A certain advantage of (a) is that it does not require ground truth or ground truth approximation to determine whether some bins are under or over-assigned. While such an imbalance certainly signals a calibration error, the histogram lack information regarding the composition of each bin. In particular, they provide no information on whether the positive class occurrences increase according to the value of the mean predicted probability bin. This can only be measured in (b), comparing all cases against an ideal calibration histogram. For multiclass problems, each class could be compared against such a histogram, and the resulting scores averaged.

\begin{figure*}[!ht]
\centering
\includegraphics[width=0.9\textwidth]{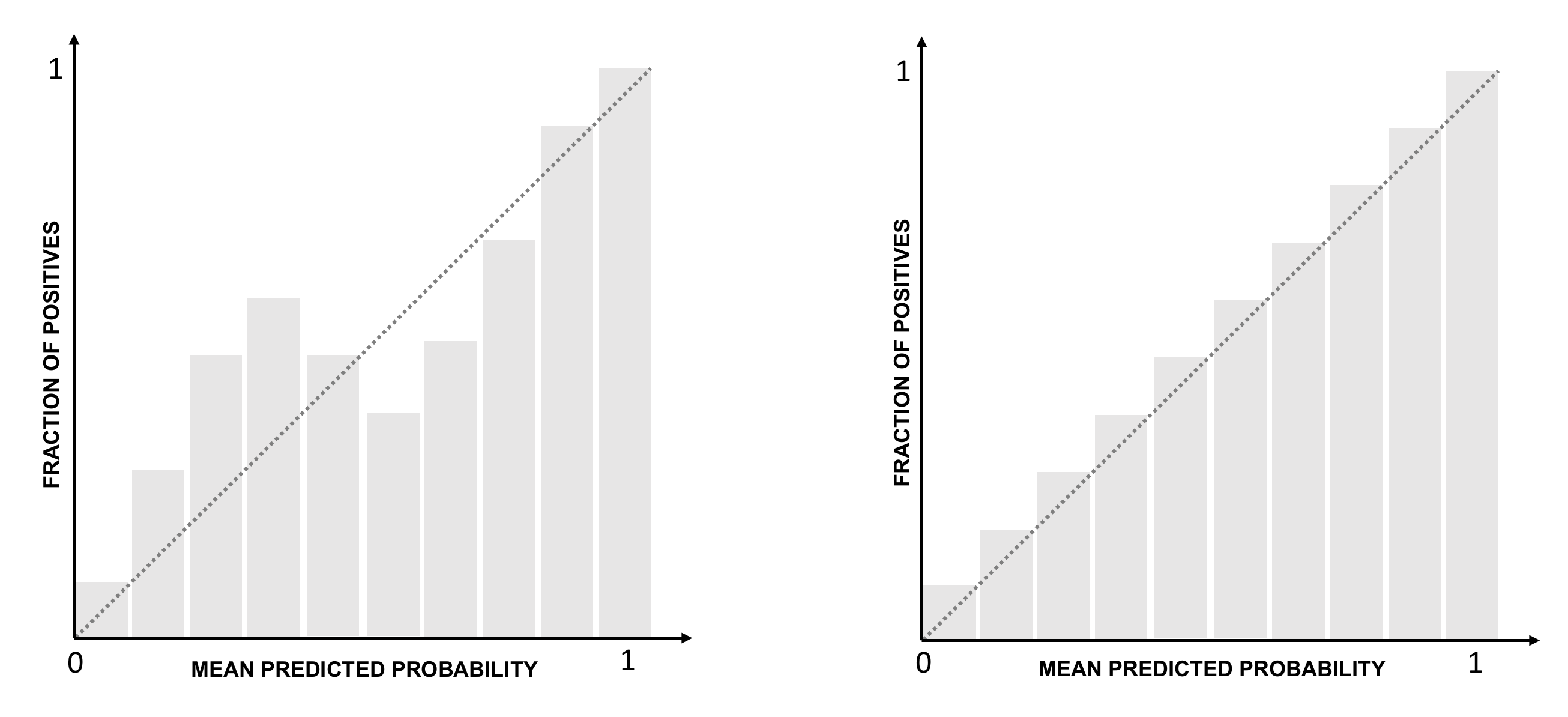}
\caption{The figure illustrates two sample histograms: the histogram on the left corresponds to some sub-optimally calibrated classifier. In contrast, the histogram on the right (reference histogram) corresponds to a perfectly calibrated classifier. The Wasserstein distance between both histograms is the minimum distance between the existing calibration and a perfect one. The distance denotes the improvement opportunity regarding the specific calibration model to achieve a perfect calibration.}
\label{F:CALIB-HISTOGRAM-DISTANCE}
\end{figure*}

To estimate how close a probability calibration method is w.r.t. the target (ideal) histogram, optimal transport is used \citep{villani2009optimal,peyre2019computational}. In particular, the Wasserstein distance between two histogram distributions is considered: a histogram constructed with the calibrator scores and a histogram corresponding to the ideal scenario. Based on them, we propose a metric that can be used to estimate the quality of calibration of any calibrator given certain ground truth. We name it \textit{Probability Calibration Score} (PCS - see Eq. \ref{E:PCS}). The proposed metrics issue a value between zero and one: PCS is zero when the model is not calibrated and one when the model is perfectly calibrated. Furthermore, a weighted metric variant can also be considered (wPCS - see Eq. \ref{E:wPCS}), where the proportion of each class among the observed instances weights the Wasserstein distances.

\begin{eqfloat}
\begin{equation}\label{E:PCS}
    PCS = \sum_{i=1}^{n} \frac{1 - W_{1}(h_{i}, h_{ref})}{n} 
\end{equation}
\caption{$W_{1}(h_{i}, h_{ref})$ is the $1$-Wasserstein distance between the histogram $h_{i}$ and the reference histogram $h_{ref}$ and $n$ is the number of classes.}
\end{eqfloat}

\begin{eqfloat}
\begin{equation}\label{E:wPCS}
    wPCS = \sum_{i=1}^{n} \left(1 - W_{1}(h_{i}, h_{ref})\right) \cdot w_{i}
\end{equation}
\caption{$W_{1}(h_{i}, h_{ref})$ is the $1$-Wasserstein distance between the histogram $h_{i}$ and the reference histogram $h_{ref}$ and $w_{i}$ is the weight of a particular class. $n$ indicates the number of classes under consideration.}
\end{eqfloat}

\begin{figure*}[!ht]
\centering
\includegraphics[width=0.9\textwidth]{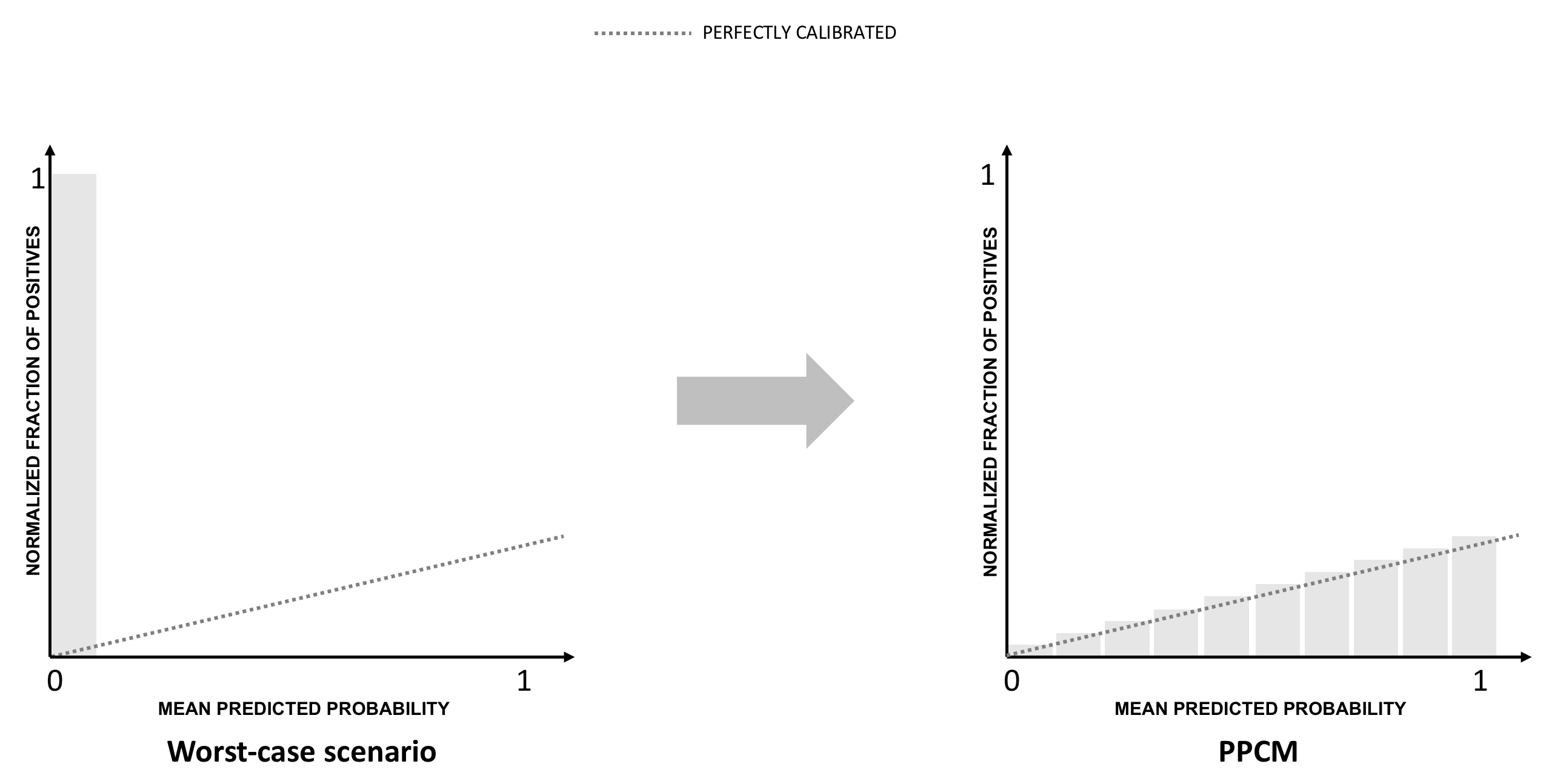}
\caption{The figure illustrates two sample calibration plots: the calibration plot on the left corresponds to a calibrated classifier where all positives were assigned to a zero mean predicted probability (worst-case scenario). In contrast, the calibration plot on the right (reference histogram) corresponds to a perfectly calibrated classifier. Both calibration plots correspond to normalized cases, where the sum of the values equals one. The Wasserstein distance between the case on the left and the distribution of a Perfect Probability Calibration Model (PPCM) is the highest among possible calibration scenarios.}
\label{F:CALIB-HISTOGRAM-MAX-DISTANCE}
\end{figure*}

To ensure the histograms are comparable, they are normalized ensuring that the sum of their values equals one. To ensure the Wasserstein distance remains between zero and one, the distance between both distributions is divided by the distance measured between the worst-case scenario and the reference ideal histogram (see Fig. \ref{F:CALIB-HISTOGRAM-MAX-DISTANCE})

When assessing the performance of an approximate calibrated model, two errors must be taken into account: (i) the classification error, given the classifier does not perfectly predict the target class (and the ground truth is approximated with such predictions), and (ii) the probability calibration technique, which does not produce a perfect probabilistic mapping between the predicted scores and the (approximated) target class. To measure (i), we choose the AUC ROC metric, which is not affected by the class imbalance. AUC ROC can be computed in a multiclass setting with a one-vs-rest or one-vs-one strategy. We measure it on the test set. We consider (ii) can be measured using the Wasserstein distance, comparing the ideal calibration histogram and a histogram where the proportion of positive class occurrences (given the approximate ground truth) is considered per mean predicted probability bin.

We propose two metrics, which we name Additive Probability Calibration Score (APCS - see Eq. \ref{E:APCS}) and Multiplicative Probability Calibration Score (MPCS - see Eq. \ref{E:MPCS}). Both summarize the calibrated models' performance, considering the classifier's imperfection (see Eq. \ref{E:KPCS}) and the calibration error incurred due to the lack of ground truth. To ensure the Wasserstein distance remains between zero and one, we compute a normalized histogram, ensuring the area of the entire histogram equals one. The proposed metrics issue a value between zero and one, and in both cases, the higher the value, the better the model. Furthermore, we also provide a weighted version of both metrics (wAPCS (see Eq. \ref{E:wAPCS}) and wMPCS (see Eq. \ref{E:wMPCS})), which aim to weight the Wasserstein distance between the normalized histograms obtained from a calibrator and the ideal histogram with the class weights (see Eq. \ref{E:APCS-W} and Eq. \ref{E:wAPCS-W} for APCS\textsubscript{W} and wAPCS\textsubscript{W}, and Eq. \ref{E:PCS} and Eq. \ref{E:wPCS} for MPCS and wMPCS).

APCS is zero when the model has no discriminative power and is not calibrated, and one when the model is perfectly calibrated and shows no classification error on the test set. The APCS metric is detailed in Eq. \ref{E:APCS}.

\begin{eqfloat}
\begin{equation}\label{E:KPCS}
    K_{AUC ROC} = \lvert0.5 - AUC ROC_{Classifier_{test}}\rvert
\end{equation}
\caption{$K$ is used to measure classifiers' discriminative power. $AUC ROC\textsubscript{Classifier\textsubscript{test}}$ corresponds to the classifiers' AUC ROC measured on the test set.}
\end{eqfloat}

\begin{eqfloat}
\begin{equation}\label{E:APCS-W}
    APCS_{W} = 0.5 \cdot PCS
\end{equation}
\caption{Component for Wasserstein distance measurement between an ideal calibrator and the calibrator under consideration, as used for the APCS metric.}
\end{eqfloat}

\begin{eqfloat}
\begin{equation}\label{E:wAPCS-W}
    wAPCS_{W} = 0.5 \cdot wPCS
\end{equation}
\caption{Component for Wasserstein distance measurement between an ideal calibrator and the calibrator under consideration, as used for the wAPCS metric.}
\end{eqfloat}

\begin{eqfloat}
\begin{equation}\label{E:APCS}
    APCS = K_{AUC ROC} + APCS_{W} 
\end{equation}
\caption{APCS metric definition.}
\end{eqfloat}

\begin{eqfloat}
\begin{equation}\label{E:wAPCS}
    wAPCS = K_{AUC ROC} + wAPCS_{W} 
\end{equation}
\caption{wAPCS metric definition.}
\end{eqfloat}

On the other hand, MPCS and wMPCS correspond to zero when (a) the classifiers' predictive ability is no better than random guessing or (b) the Wasserstein distance between histograms is highest (equal to one). Moreover, MPCS and wMPCS correspond to one when (a) the classifiers' predictive ability is perfect, and (b) the calibration is perfect w.r.t. the target histogram \textit{h} of choice. The MPCS metric is detailed in Eq. \ref{E:MPCS}.

\begin{eqfloat}
\begin{equation}\label{E:MPCS}
    MPCS = K_{AUC ROC} \cdot PCS 
\end{equation}
\caption{MPCS metric definition}
\end{eqfloat}

\begin{eqfloat}
\begin{equation}\label{E:wMPCS}
    wMPCS = K_{AUC ROC} \cdot wPCS 
\end{equation}
\caption{MPCS metric definition}
\end{eqfloat}

\begin{figure*}[!ht]
\centering
\includegraphics[width=0.9\textwidth]{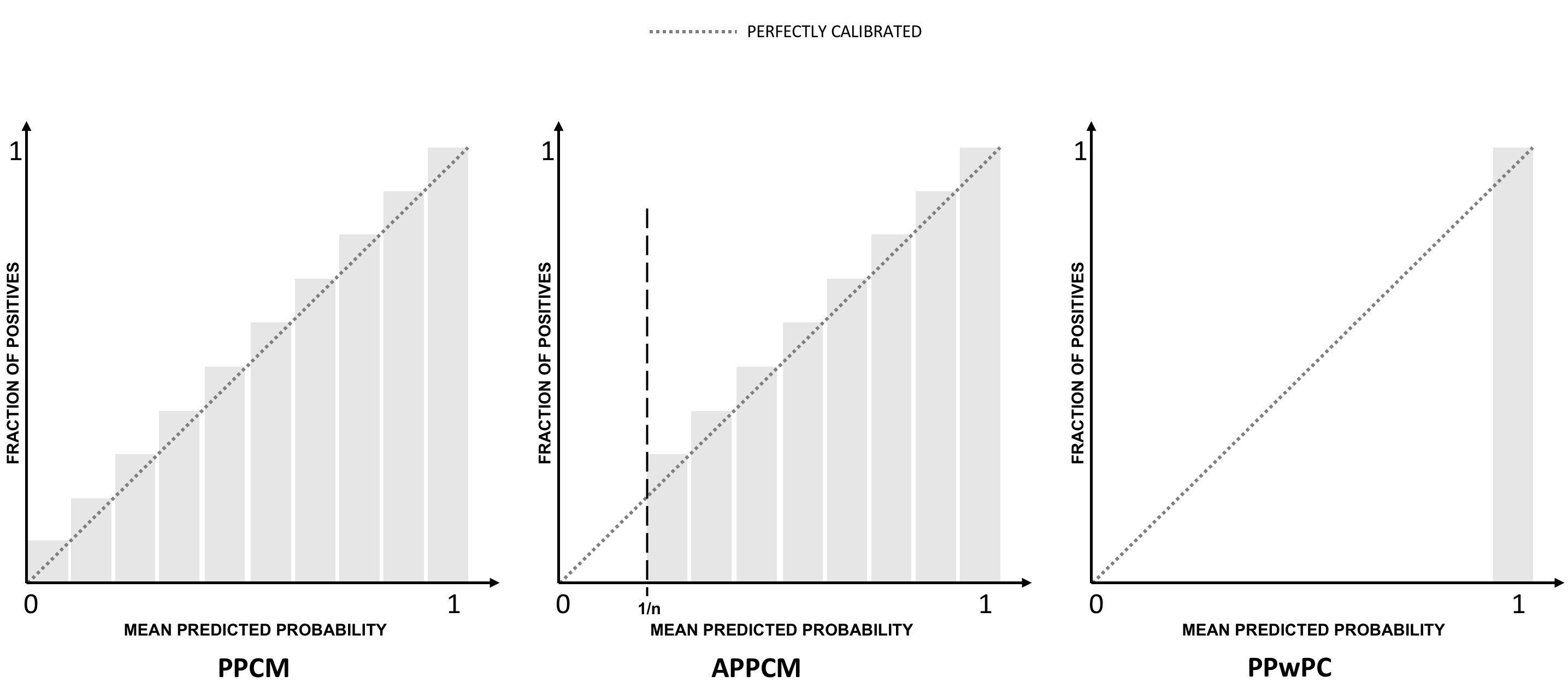}
\caption{The figure illustrates three histograms that correspond to ideal cases described in this section: Perfect Probability Calibration Model (PPCM), Almost Perfect Probability Calibration Model (APPCM), and Perfect Classification with Perfect Confidence (PPwPC).}
\label{F:CALIB-HISTOGRAM-PERFECT-CASES}
\end{figure*}

For models' probability calibration, PCS, APCS, and MPCS assume an ideal reference histogram. Three histograms are presented in Fig. \ref{F:CALIB-HISTOGRAM-PERFECT-CASES} corresponding to (a) a Perfect Probability Calibration Model (PPCM), (b) an Almost Perfect Probability Calibration Model (APPCM), and (c) Perfect Classification with Perfect Confidence (PPwPC). While only PPCM can be used for strict probability calibration, the other two reference histograms measure how far the distributions of the predicted values are from other desired distribution shapes. In particular, APPCM achieves a similar spread of predicted probabilities as PPCM but neglects the segment of predictions below $1/n$ (with $n=number \, of \, classes$), where the classifier would become suboptimal. On the other hand, PPwPC advocates for a classifier where all scores are pushed toward the highest possible score for a given class. This research only considers the PPCM reference histogram to compute the above-described metrics.

\section{Use case}\label{S:USE-CASE}
\begin{figure*}[ht]
\centering
\includegraphics[width=0.70\textwidth]{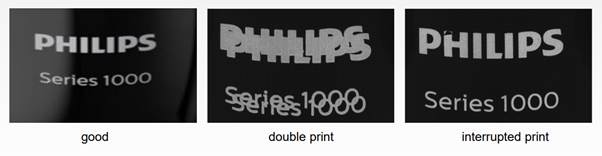}
\caption{The images shown above correspond to three possible classes: good (no defect), double print (defective), and interrupted print (defective).}
\label{F:PHIA-CLASSES}
\end{figure*}

\textit{Philips Consumer Lifestyle BV} in Drachten, The Netherlands, is one of Philips' biggest development and production centers in Europe. They use cutting-edge production technology to manufacture products ceaselessly. One of their improvement opportunities is related to visual inspection, where they aim to identify when the company logo is not properly printed on the manufactured products. They have multiple printing pad machines, from which the products are handled and inspected on their visual quality and removed if any error is detected. Experts estimate that a fully automated procedure would speed up the process by more than 40\%. Currently, there are two defects associated with the printing quality of the logo (see Fig. \ref{F:PHIA-CLASSES}): double prints (the whole logo is printed twice with a varying overlap degree) and interrupted prints (the logo displays small non-pigmented areas, similar to scratches).

Machine learning models can be developed to automate the visual inspection procedure \citep{rippel2021anomaly,zavrtanik2022dsr}. However, given that such models are imperfect, the manual revision can be used as a fallback to inspect the products about which the uncertainty of the machine learning model exceeds a certain threshold. Such decisions can be made based on simple decision rules, quality policies, and the probability of obtaining a defective product given a particular prediction score. Furthermore, products sent for manual inspection can be prioritized using different criteria to enhance the existing defect detection machine learning model. This research explores the abovementioned capabilities through multiple experiments, building supervised models, leveraging active learning, and comparing six machine learning algorithms. Furthermore, new measures for probability calibration are explored, and experiments are executed to determine whether existing calibration techniques would benefit from enlarging the calibration set with approximate ground truth. The experiments were conducted on a dataset of 3518 labeled images, all corresponding to manufactured shavers.

\section{Methodology}\label{S:METHODOLOGY}

The research presented in this paper was performed using the Python language, and open source libraries, such as scikit-learn \citep{sklearn_api} and netcal \citep{Kueppers_2020_CVPR_Workshops}.

\subsection{Methodological Aspects to Evaluate Active Learning Strategies}

\begin{figure*}[ht]
\centering
\includegraphics[width=0.90\textwidth]{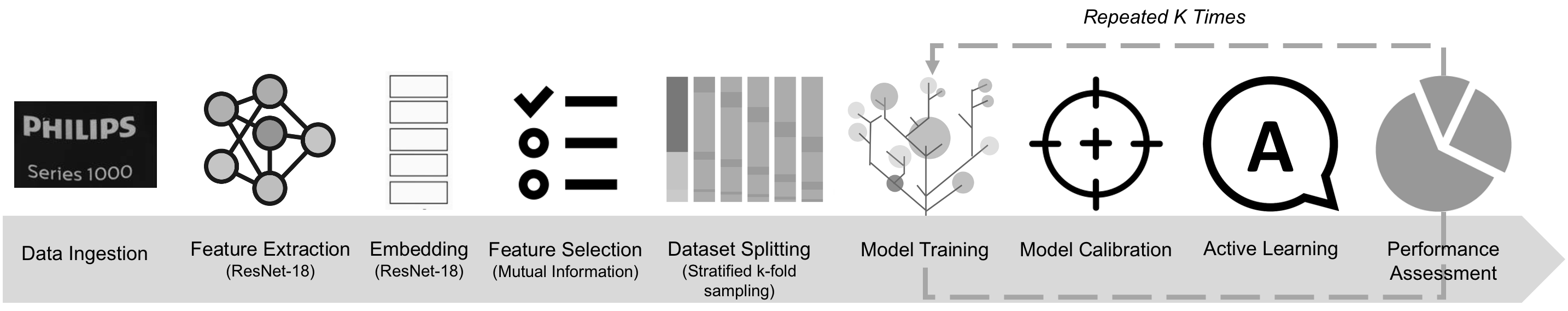}
\caption{The methodology we followed to train and assess machine learning models and active learning scenarios.}
\label{F:METHODOLOGY}
\end{figure*}

We frame the automated defect detection as a supervised, multiclass classification problem. A ResNet-18 model \citep{he2016deep} was used for feature extraction. 512 values long vectors were extracted for each image obtained from the average pooling layer. To avoid overfitting, the procedure suggested by \cite{hua2005optimal} was followed by selecting the \textit{top K} features, with $K=\sqrt{N}$, where N is the number of data instances in the train set. Features' relevance was assessed considering the mutual information score, which measures any relationship between random variables. It is considered that the mutual information score is not sensitive to feature transformations if these transformations are invertible and differentiable in the feature space or preserve the order of the original elements of the feature vectors \citep{vergara2014review}.

\begin{figure*}[ht]
\centering
\includegraphics[width=0.70\textwidth]{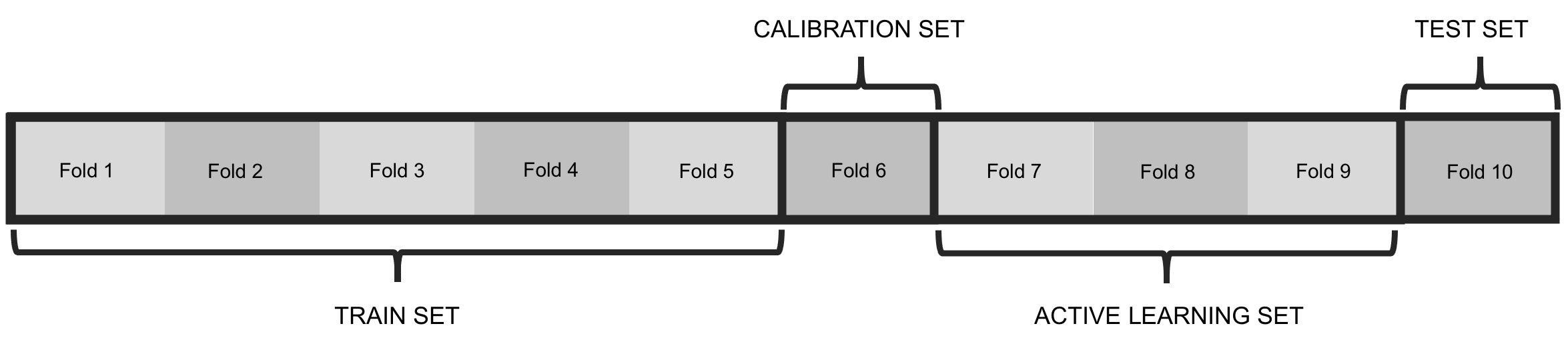}
\caption{A ten-fold stratified cross-validation was used. The dataset was split for four purposes: train, test, probabilities calibration, and simulate unlabeled data under an active learning setting.}
\label{F:KFOLDS}
\end{figure*}

To evaluate the models' and active learning scenarios' performance, a stratified k-fold cross validation \citep{zeng2000distribution} was applied, considering \textit{k=10} based on recommendations by \cite{kuhn2013applied}. One fold was used for testing (\textit{test set}), and one for machine learning models' probabilities calibration (\textit{calibration set}). Three folds were used to simulate a pool of unlabeled data for active learning (\textit{active learning set}), and the rest to train the model (\textit{train set}) (see Fig. \ref{F:KFOLDS}). Samples are selected from the \textit{active learning set} to be annotated by the oracle and then added to the training set, on which the models are retrained. In this research, two types of oracles were considered: (a) machine oracles, which can be imperfect, and (b) human annotators (assumed to be ideal). Five machine learning algorithms were evaluated: Gaussian N\"aive Bayes, CART (\textit{Classification and Regression Trees}, similar to C4.5, but it does not compute rule sets), Linear SVM, kNN, and Multilayer perceptron (MLP).

To evaluate the discriminative power of the machine learning models and how it is enhanced over time through active learning, the AUC ROC metric was computed. Given the multiclass setting, the "one-vs-rest" heuristic was selected, splitting the multiclass dataset into multiple binary classification problems and computing their average, weighted by the number of true instances for each class. In addition, to assess the usefulness of the active learning approaches, the AUC ROC values obtained by evaluating the model against the test fold for the first (Q1) and last (Q4) quartiles of instances queried in an active learning setting were compared. The amount of manual work saved under each active learning setting and the soft-labeling approaches' precision were also evaluated.

\begin{figure*}[ht]
\centering
\includegraphics[width=0.70\textwidth]{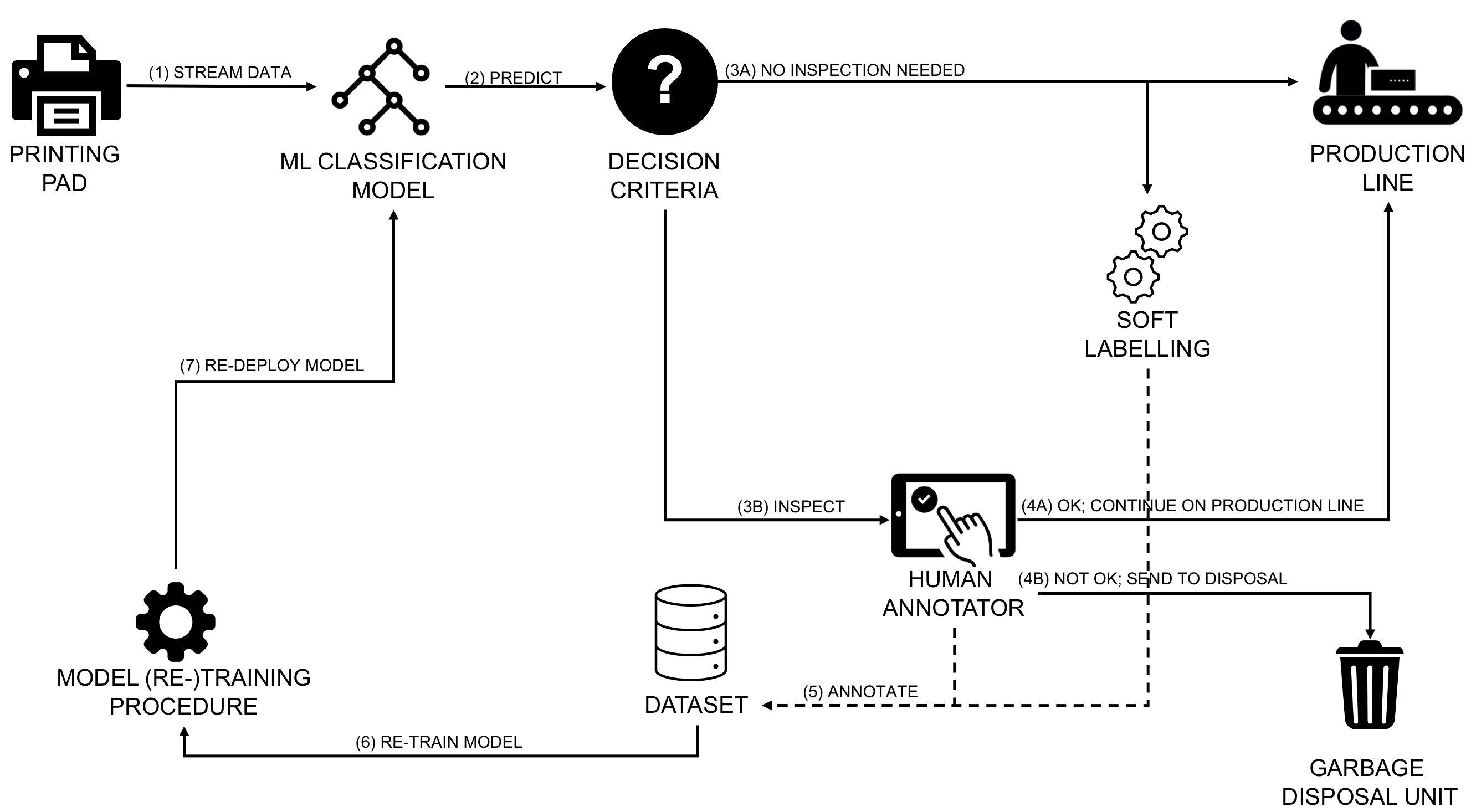}
\caption{Expected visual inspection pipeline in a production setting. Multiple active learning strategies were assessed to identify which would drive the best results.}
\label{F:PIPELINE}
\end{figure*}

Through different experiments (detailed in Section \ref{S:EXPERIMENTS}), a visual inspection pipeline was simulated (see Fig. \ref{F:PIPELINE}). First, a stream of images is directed toward the machine learning model trained to identify possible defects. Then, based on the prediction score, a decision is made on whether the manufactured product should remain in the production line or be deferred to manual inspection. If the product is unlikely to be defective, such a decision can be considered a label (it is considered a soft label when not made by a human annotator). The label is then persisted, adding enlarging the existing dataset. The enlarged dataset can be used to retrain the model and replace the existing one after a successful deployment.

\subsection{Methodological Aspects to Evaluate Probability Calibration Metrics and Strategies}

\begin{figure*}[ht]
\centering
\includegraphics[width=0.70\textwidth]{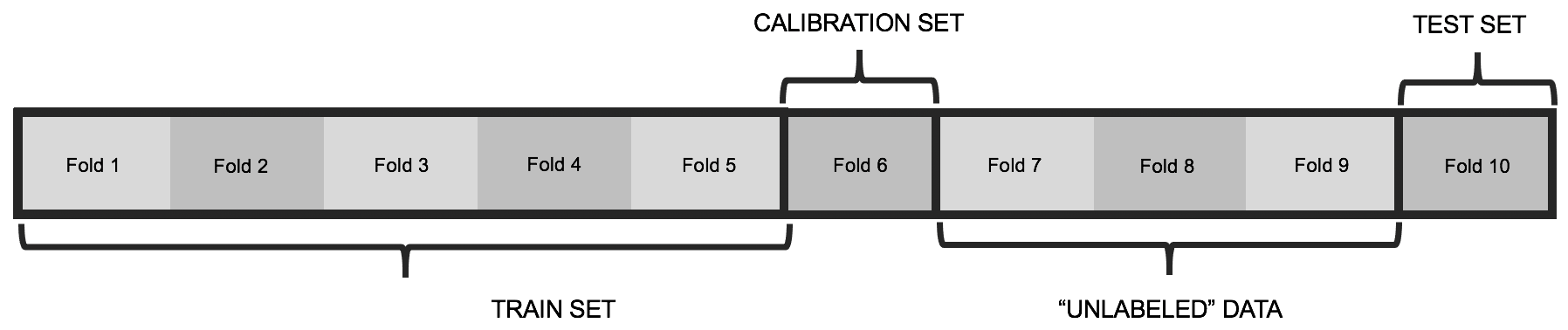}
\caption{A ten-fold stratified cross-validation was used. The data was split into train set, test set, calibration set, and \textit{unlabeled data}. \textit{Unlabeled data} was used to simulate a stream of unlabeled data and assess whether a histogram-based calibration method without ground truth can enhance its performance over time.}
\label{F:KFOLDS-CALIBRATION}
\end{figure*}

\begin{eqfloat}
\begin{equation}\label{E:wECE}
    wECE = \sum_{i=1}^{n} ECE_{i} \cdot w_{i}
\end{equation}
\caption{wECE metric definition. $n$ indicates the number of classes under consideration.}
\end{eqfloat}

To evaluate the proposed probability calibration metrics and techniques, a similar procedure as the one described in the previous subsection was followed, avoiding the active learning step. Furthermore, a different dataset split was considered (see Fig. \ref{F:KFOLDS-CALIBRATION}). After training the machine learning model and calibrating it with the calibration set, the non-calibrated model was used to issue a prediction for each instance in the \textit{unlabeled data} set. The predicted class is then used to adjust further (train) the calibrator. While this introduces some noise, we expect that the better the classification model, the more it would benefit the calibrator, as explained in Section \ref{S:PROBABILITIES-CALIBRATION}. Eleven performance metrics were measured: AUC ROC, ECE, wECE (see Eq. \ref{E:wECE}), PCS, wPCS, APCS\textsubscript{W}, wAPCS\textsubscript{W}, APCS, wAPCS, MPCS, and wMPCS. AUC ROC measures the discriminative capability of the model and provides insights into how such capability is affected by different calibration techniques. ECE evaluates the expected difference between the accuracy and confidence of a calibration model. The ECE metric was used to compare the calibration quality for the multiple calibration techniques and the newly proposed PCS, wPCS, APCS, wAPCS, MPCS, and wMPCS metrics. Furthermore, given that the newly proposed metrics were built on a similar concept as the ECE metric, we are interested in how much they capture the same information. The Kendall $\tau$ (see \citep{kendall1938new}) and the Pearson correlation between ECE and the newly proposed metrics were measured. The Kendall correlation measures the ordinal association between two measured quantities. In this case, it measures to what extent both metrics increase or decrease, given the predictions for a given machine learning model and calibrators. The Pearson correlation, on the other side, was used to assess whether the correlation between metrics was linear.

The metrics were computed on the test set against the ground truth (class annotations) and the approximate ground truth (predicted classes). The results were analyzed to understand how well the metrics capture the models' performance and calibration when no ground truth is available. Furthermore, the weighted and non-weighted metrics were compared to understand how class weighting influences the final score and perception regarding the quality of the calibration.

\section{Experiments}\label{S:EXPERIMENTS}
\subsection{Experimenting with Active Learning Strategies}

\begin{figure*}[!ht]
\centering
\includegraphics[width=0.9\textwidth]{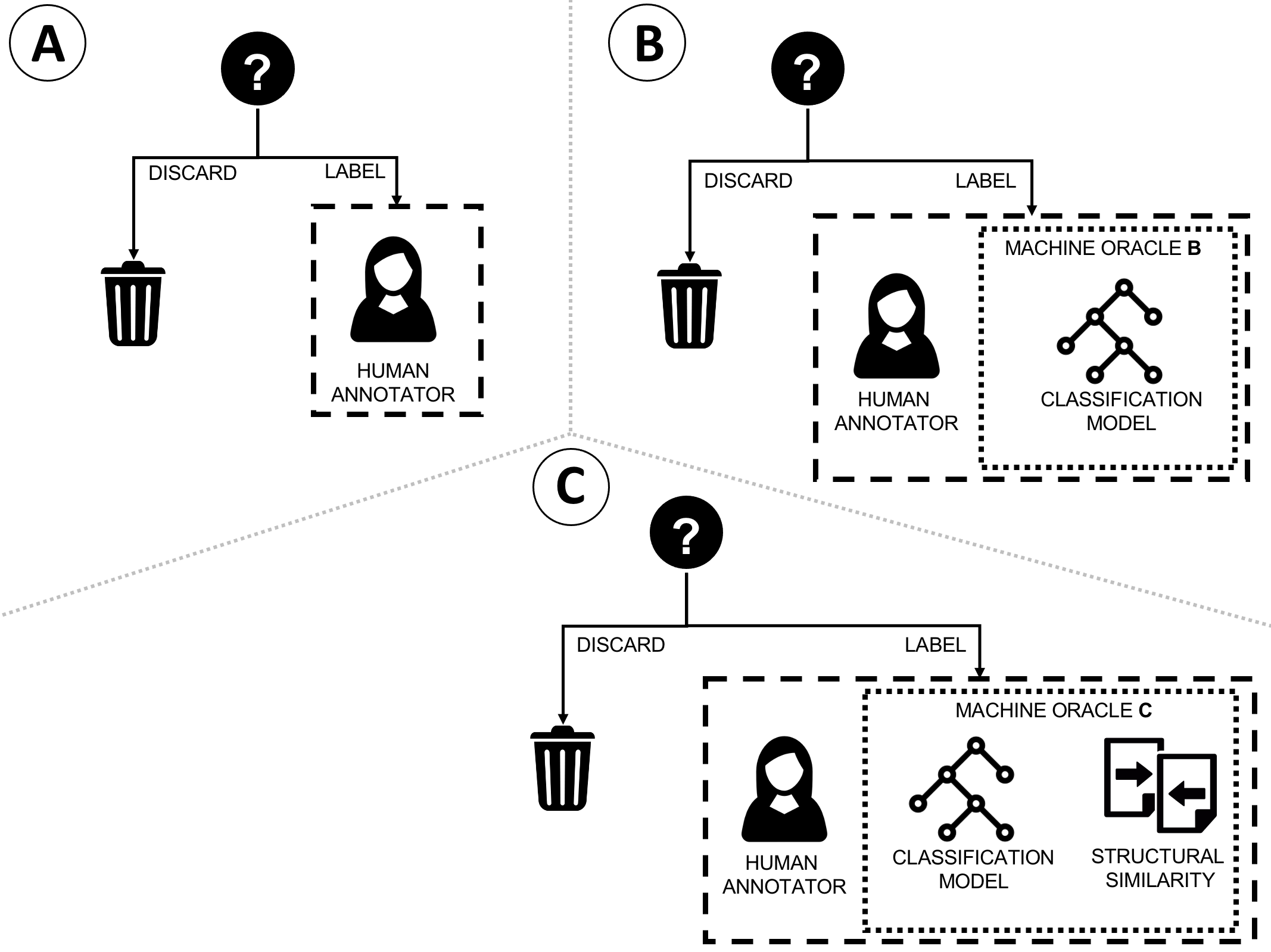}
\caption{Three oracle settings are explored in this research: (A) human annotator, (B) soft-labeling with classification model's outcomes for instances with high-confidence scores, and human annotator for instances where the model has low confidence; and (C) which is analogous to (B), but the machine oracle takes into account the classifier's output score and whether the predicted class matches the class with of a labeled image with the shortest distance towards the active sample. In (C), the sample is sent to manual revision if there is a class mismatch in the machine oracle. Samples are only discarded in a streaming setting.}
\label{F:ORACLES}
\end{figure*}

For this research, two active learning settings were explored (pool-based and stream-based), using four distinct strategies to label the queried data instances in an active learning setting. Two strategies were used to select data from the active learning set under the pool-based active learning setting: (a) random sampling and (b) instances for which the classification model was the most uncertain. The model's uncertainty was assessed by considering the highest score for a given class for a given instance and selecting the instance with the lowest score among the scores provided for the data instances in the active learning set. In both cases, data were sampled until the set's exhaustion. Under the streaming active learning setting, a slightly different policy was used. When random sampling was used, a decision was made whether to keep or discard the instance with a probability threshold of 0.5. Under the highest uncertainty selection criteria, the prediction for each data instance was analyzed and derived to the oracles for labeling if it was below a certain confidence threshold (p=0.95 or p=0.99).

Three oracle settings were considered (see Fig. \ref{F:ORACLES}): (A) human labeler as the only source of truth, (B) machine oracle (classifier model) for data instances where the classifier had a high certainty, and a human labeler otherwise; and (C) machine oracle (classifier model) for data instances where the classifier had a high certainty, and requesting an additional \textit{opinion} to another machine oracle when uncertain about the outcome. This second oracle queries the closest labeled image from three randomly picked images (one per class). In (C), the machine oracle issues a label only when both machine oracles are unanimous on the label; otherwise, the instance labeling is delegated to a human labeler. The decision regarding which oracle to query was based on the models' confidence regarding the outcome and a probability threshold set based on manufacturing quality policies. It was assumed that the second machine oracle in (C) is accessible at a certain cost (e.g., paid external service) and, therefore, cannot be used for every prediction. Such a service was simulated by computing the Structural Similarity Index Measure (SSIM) score over the queried image.

Eight scenarios were set up (see Table \ref{T:EXPERIMENTS}), and experimented with two quality thresholds (0.95 and 0.99 probability that the item corresponded to a certain class) and five machine learning models. The machine learning models were calibrated using a sigmoid model based on Platt logistic model \citep{platt1999probabilistic} (see Eq. \ref{E:PLATT-CALIBRATION}).

\begin{eqfloat}
\begin{equation}\label{E:PLATT-CALIBRATION}
P(y_i=1 \mid f_i) = \frac{1}{1+exp(Af_i+B)}
\end{equation}
\caption{Platt classifier calibration logistic model. $y_i$ denotes the truth label, and $f_i$ denotes the uncalibrated classifier's prediction for a particular sample. $A$ and $B$ denote adjusted parameters when fitting the regressor.}
\end{eqfloat}

\begin{table*}[ht!]
\resizebox{\textwidth}{!}{
\begin{tabular}{|l|l|l|l|}
\hline
\textbf{Experiment} & \textbf{AL setting} & \textbf{AL data selection} & \textbf{Oracle} \\ \hline
1 & pool-based & Random sampling & Human labeler \\ \hline
2 & pool-based & Highest uncertainty & Human labeler \\ \hline
3 & pool-based & Highest uncertainty & Machine Oracle B + Human labeler \\ \hline
4 & pool-based & Highest uncertainty & Machine Oracle C + Human labeler \\ \hline
5 & stream-based & Random sampling & Human labeler \\ \hline
6 & stream-based & Highest uncertainty & Human labeler \\ \hline
7 & stream-based & Highest uncertainty & Machine Oracle B + Human labeler \\ \hline
8 & stream-based & Highest uncertainty & Machine Oracle C + Human labeler \\ \hline
\end{tabular}
\caption{Proposed experiments to evaluate the best active learning setting regarding how it influences the models' learning and its impact on the manual revision workload. \label{T:EXPERIMENTS}}}
\end{table*}

\subsection{Experiments Assessing Probability Calibration Metrics and Techniques}
In an automated visual inspection setting, a labeling effort is required to (a) label data to train and calibrate the machine learning models and (b) perform a manual inspection when the models cannot determine the class of a given data instance accurately. To understand how the probability calibration affects the machine learning models, the models' predictions were compared against those obtained by (a) not calibrating the model (\textit{No calibration}) and calibrating the model with (b) a sigmoid model based on the Platt logistic model (\textit{Platt}), (c) temperature scaling (\textit{Temperature}), and (d) \textit{Histogram} calibration. Two aspects were considered in the experiments: (i) how calibration techniques compare against each other and (ii) whether calibrating a model without a ground truth can provide comparable results to models calibrated with ground truth.

\section{Results and Evaluation}\label{S:RESULTS-AND-EVALUATION}
\subsection{Results and Evaluation of Active Learning Strategies}

The active learning strategies were analyzed from two points of view. First, whether they contributed to better learning of the machine learning model. Second, how much manual work could be saved by adopting such strategies.

\begin{table*}[ht!]
\resizebox{\textwidth}{!}{
\begin{tabular}{|c|l|ll|ll|}
\hline
\multirow{2}{*}{\textbf{Setup}} & \multicolumn{1}{c|}{\multirow{2}{*}{\textbf{Experiment}}} & \multicolumn{2}{c|}{\textbf{p=0.95}} & \multicolumn{2}{c|}{\textbf{p=0.99}} \\ \cline{3-6} 
 & \multicolumn{1}{c|}{} & \multicolumn{1}{l|}{\textbf{Q1}} & \textbf{Q4} & \multicolumn{1}{l|}{\textbf{Q1}} & \textbf{Q4} \\ \hline
\multirow{4}{*}{\textbf{Pool-based}} & 1 - random, human oracle & \multicolumn{1}{l|}{0.8428} & 0.8612 & \multicolumn{1}{l|}{0.8431} & 0.8623 \\ \cline{2-6} 
 & 2 - uncertainty, human oracle & \multicolumn{1}{l|}{\textbf{0.8594}} & \textbf{0.8693} & \multicolumn{1}{l|}{\textbf{0.8594}} & \textbf{0.8693} \\ \cline{2-6} 
 & 3 - uncertainty, oracle (machine B + human) & \multicolumn{1}{l|}{0.8398} & 0.8396 & \multicolumn{1}{l|}{0.8398} & 0.8398 \\ \cline{2-6} 
 & 4 - uncertainty, oracle (machine C + human) & \multicolumn{1}{l|}{0.8349} & 0.8348 & \multicolumn{1}{l|}{0.8358} & 0.8358 \\ \hline
\multirow{4}{*}{\textbf{Streaming}} & 5 - random, human oracle & \multicolumn{1}{l|}{0.8460} & 0.8559 & \multicolumn{1}{l|}{0.8460} & 0.8559 \\ \cline{2-6} 
 & 6 - uncertainty, human oracle & \multicolumn{1}{l|}{0.8525} & 0.8647 & \multicolumn{1}{l|}{0.8529} & 0.8647 \\ \cline{2-6} 
 & 7 - uncertainty, oracle (machine B + human) & \multicolumn{1}{l|}{0.8505} & 0.8608 & \multicolumn{1}{l|}{0.8529} & 0.8647 \\ \cline{2-6} 
 & 8 - uncertainty, oracle (machine C + human) & \multicolumn{1}{l|}{\textit{0.8550}} & \textit{0.8665} & \multicolumn{1}{l|}{\textit{0.8553}} & \textit{0.8668} \\ \hline
\end{tabular}
\caption{Mean values for the mean AUC ROC computed across ten folds for five machine learning models. The results show how different active learning policies influence the models' learning over time (Q1 (first quartile) vs. Q4 (last quartile)). Two probability thresholds (0.95 and 0.99) were considered as a soft labeling cut-off. The best results are bolded, and the second-best ones are displayed in italics. \label{T:AUC-ROC-VALUES-ALL}}}
\end{table*}

For the first case, the AUC ROC was measured over time (see Table \ref{T:AUC-ROC-VALUES-ALL}). In particular, the models' average performance was contrasted when they consumed data within the Q1 and Q4 of the active learning pool. The best outcomes were observed for Experiment 2 (highest uncertainty with human labeler) settings, while the second-best performance was observed for Experiment 8 (highest uncertainty, with the machine and human oracles). Overall, it was observed that the streaming setting had a better average performance when compared to the pool-based experiments, despite achieving only the second-best results with Platt scaling. Furthermore, in two cases, the machine learning model degraded its performance between Q1 and Q4. This happened for Experiment 3 ($p=0.95$) and Experiment 4 ($p=0.95$). 

Given that (a) in both experiments, a machine oracle was used, (b) no performance decrease was observed for $p=0.99$, and (c) that the same setting did not affect the streaming case, we were tempted to conclude that most likely the machine oracles mislabeled certain instances, confusing the model when retrained and therefore reducing the model's performance over time. Nevertheless, further analysis revealed a small fraction of soft-labeled data and that most cases were accurately labeled. While soft labeling was detrimental for the pool-based active learning settings, it led to superior results in a streaming setting, achieving results close to the best ones obtained across all experiments.

\begin{table*}[ht!]
\resizebox{\textwidth}{!}{
\begin{tabular}{|l|l|l|l|}
\hline
\textbf{Model} & \textbf{Q1} & \textbf{Q4} & \textbf{DS(p=0,95)} \\ \hline
MLP & \textbf{0.9309$\pm$0.0004} & \textbf{0.9448$\pm$0.0003} & Yes \\ \hline
SVM & \textit{0.8788$\pm$0.0007} & \textit{0.8767$\pm$0.0007} & Yes \\ \hline
NB & 0.8628$\pm$0.0005 & 0.8675$\pm$0.0005 & Yes \\ \hline
KNN & 0.8575$\pm$0.0006 & 0.8720$\pm$0.0006 & Yes \\ \hline
CART & 0.7669$\pm$0.0007 & 0.7854$\pm$0.0008 & Yes \\ \hline
\end{tabular}
\caption{Mean AUC ROC values computed across ten test folds for five machine learning models. The results show how the machine learning models learn over time (Q1 vs. Q4) under the Experiment 2 setting. Furthermore, we analyze if the differences were statistically significant at a p-value=0.95 (DS(p=0.95)). The best results are bolded, and the second-best results are displayed in italics. \label{T:AUC-ROC-VALUES-EXP2}}}
\end{table*}

In Table \ref{T:AUC-ROC-VALUES-EXP2} we report the performance of machine learning models for Experiment 2 and compare how they performed after Q1 and Q4 of the active learning pool data was shown to them. We found that the best performance was attained by the MLP, followed by the SVM by at least 0.05 AUC ROC points. Furthermore, while the MLP increased its performance over time, the SVM slightly reduced it in Q4. No other model had a performance decrease over time. Since Experiment 2 only considered a human oracle and the annotations are accurate, the performance decrease cannot be attributed to mislabeling. Furthermore, while the model's discriminative capacity loss could be attributed to the class imbalance, we consider this improbable, given that the rest of the models could better discern among the classes over time. Finally, the CART model obtained the worst results, which lagged slightly more than 0.16 AUC ROC points compared to the best one. 

As mentioned at the beginning of this section, another relevant aspect of evaluating active learning strategies is their potential to reduce data annotation efforts. This could be analyzed from two perspectives. First, whether the additional data annotations provide enough knowledge to enhance the models' performance significantly. If not, the data annotation can be avoided. Second, a strategy can be devised (e.g., a machine oracle) to reduce the manual annotation effort. In this work, we focused on the second one. Table \ref{T:LABELING-SAVINGS-P95} presents the results for a cut-off value of p=0.95. For p=0.99, no instances were retrieved and given to machine oracles; therefore, no analysis was performed on them.

When considering the cut-off value of 0.95, it was noticed that the Platt calibration considered a negligible number of cases for each experiment. While the quality of the annotations was high, using machine oracles would not strongly alleviate the manual labeling effort. The highest amount of soft-labeled instances corresponded to experiments with streaming settings (Experiment 7 and Experiment 8), which soft-labeled 4\% and 3\% of all data instances, respectively. Furthermore, 96\% of samples were correctly labeled in both cases, meeting the quality threshold of p=0.95. The decrease in the amount of soft labeled samples for Experiment 8 was due to discrepancies between the machine learning model and the SSIM score. Furthermore, the best machine labeling quality was achieved when considering \textit{Oracle C} (unanimous vote of two machine oracles). When contrasting with the AUC ROC results obtained for those experiments, it was observed that while Experiments 3 and 4 slightly decreased discriminative power, Experiments 7 and 8 increased their performance for at least 0.01 AUC ROC points.

\begin{table*}[ht!]
\resizebox{\textwidth}{!}{
\begin{tabular}{|l|rrrr|}
\hline
\textbf{Experiment} & \multicolumn{4}{c|}{\textbf{p=0.95}} \\ \hline
\textbf{} & \multicolumn{1}{l|}{\textbf{SL (\%)}} & \multicolumn{1}{l|}{\textbf{SL OK (\%)}} & \multicolumn{1}{l|}{\textbf{ML SL OK (\%)}} & \multicolumn{1}{l|}{\textbf{SSIM SL OK (\%)}} \\ \hline
\textbf{3} & \multicolumn{1}{r|}{0.0077} & \multicolumn{1}{r|}{0.9684} & \multicolumn{1}{r|}{0.0075} & NA \\ \hline
\textbf{4} & \multicolumn{1}{r|}{0.0033} & \multicolumn{1}{r|}{\textbf{0.9756}} & \multicolumn{1}{r|}{0.0050} & 0.0033 \\ \hline
\textbf{7} & \multicolumn{1}{r|}{0.0413} & \multicolumn{1}{r|}{0.9685} & \multicolumn{1}{r|}{0.0400} & NA \\ \hline
\textbf{8} & \multicolumn{1}{r|}{0.0343} & \multicolumn{1}{r|}{\textit{0.9692}} & \multicolumn{1}{r|}{0.0483} & 0.0334 \\ \hline
\end{tabular}
\caption{Proportion and quality of soft labeling through different settings, considering a predicted probability cut-off value of p=0.95. The task required annotating 2460 samples on average. \textit{SL (\%)} denotes the percentage of soft annotated data instances w.r.t. the total, \textit{SL OK (\%)} denotes the percentage of correctly soft annotated instances, \textit{ML SL OK (\%)} denotes the percentage of soft annotated data instances w.r.t. the total that would be correctly annotated considering the ML model score, \textit{SSIM SL OK (\%)} denotes the percentage of soft annotated data instances w.r.t. the total that would be correctly annotated considering the SSIM score.  \label{T:LABELING-SAVINGS-P95}}}
\end{table*}

\subsection{Results and Evaluation of Probability Calibration Metrics and Techniques}

The experiments performed in this research, aimed to validate whether the metrics proposed to measure the quality of a calibrator can be used to understand the performance of a calibrator even when no ground truth is available. Furthermore, it aimed to validate whether predictions on unlabeled data could enhance the calibrators' performance. The results are presented in Table \ref{T:RESULTS-CORRELATIONS}, Table \ref{T:RESULTS-NOT-WEIGHTED}, and Table \ref{T:RESULTS-WEIGHTED}. The PCS, APCS, and MPCS (along with the weighted variants) metrics were computed considering the PPCM histogram, which denotes a perfect calibration.

To understand whether the proposed metrics can measure the calibration quality without ground truth, the Pearson and Kendall correlations were computed between the ECE, wECE, APCS\textsubscript{W}, wAPCS\textsubscript{W}, PCS, and wPCS metrics (see Table \ref{T:RESULTS-CORRELATIONS}). While ECE and wECE are always computed considering the ground truth at the test set, PCS, APCS\textsubscript{W}, wPCS, and wAPCS\textsubscript{W} were calculated considering two cases: ground truth (golden standard) and predicted labels (approximate ground truth) at the test set. Furthermore, the correlations between the metrics were evaluated in two separate moments: after calibrating the models with the calibration set (CS) and after calibrating the models with additional samples retrieved from the \textit{unlabeled data} set (CS+UD). The results show that the correlation between ECE, PCS, APCS\textsubscript{W}, wPCS, and wAPCS\textsubscript{W} metrics is consistent across all cases. Furthermore, little variation exists between the values obtained when PCS or wPCS were computed on the ground truth or the approximate ground truth. While the Pearson correlation decreases after training the calibrator with predicted labels from the unlabeled data set, the Kendall correlation grew stronger when PCS or APCS\textsubscript{W} were just averaged across classes and not weighted by the frequency of occurrence of each class. We consider the correlations moderate (Pearson correlation was measured between 0.50 and 0.61) or strong (Kendall correlation was above 0.33 and slightly below 0.40). Given the abovementioned results, we consider the PCS, wPCS, APCS\textsubscript{W}, and wAPCS\textsubscript{W} metrics adequately capture information conveyed by the ECE metric regardless of the source of truth used to measure the quality of the calibration. Therefore, we conclude that PCS, wPCS, APCS\textsubscript{W}, and wAPCS\textsubscript{W} can be used to assess the calibrators' quality when no ground truth is available.

\begin{figure*}[ht]
\centering
\includegraphics[width=\textwidth]{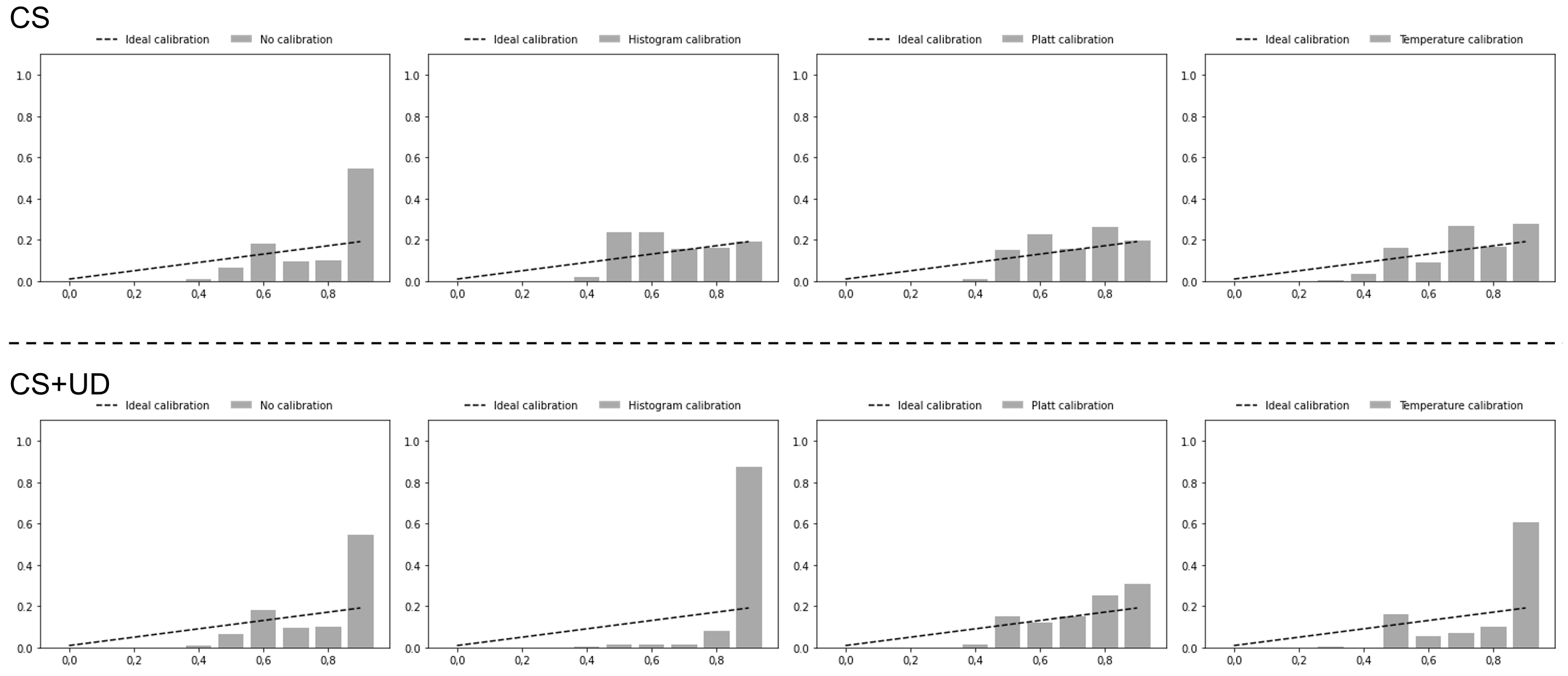}
\caption{Eight calibration plots, comparing \textit{No calibration}, \textit{Histogram calibration}, \textit{Platt calibration}, and \textit{Temperature calibration} at CS (calibrated with the calibration set (ground truth)) and CS+UD (calibrated with a calibration set and predicted labels over time). The calibration plots have been adapted, showing normalized values (their sum is one) rather than the usual fraction of positives on the dependent variable axis. The x-axis denotes the mean predicted probability for a given class. The histograms average predictions across classes and calibrated machine learning models.}
\label{F:CALIBRATION-RESULTS}
\end{figure*}

Table \ref{T:RESULTS-NOT-WEIGHTED} and Table \ref{T:RESULTS-WEIGHTED}, compare the calibrators across multiple metrics to assess how an approximate calibration affects their discriminative power (AUC ROC) and whether it helps to enhance the calibrators' quality (ECE, PCS, APCS, MPCS, and their weighted variants). Furthermore, Fig. \ref{F:CALIBRATION-RESULTS} presents calibration plots for each calibrator for CS and CS+UD for visual assessment. 

When comparing the calibrators through non-weighted metrics (see Table \ref{T:RESULTS-NOT-WEIGHTED}), we consider the Platt calibrator achieved the most stable performance. While the measured quality of calibration slightly decreased with the approximate calibration, it must be noticed that a higher proportion of positives was allocated at higher scores. Furthermore, while with the approximate calibration, the model's overall discriminative power slightly decreased, it remained superior against other models (even the not calibrated) by at least 0.02 AUC ROC points. The Histogram and Temperature calibrators provide an interesting case, given both had a similar initial (CS) calibration quality if measured with the ECE, PCS, APCS, or MPCS metrics. Nevertheless, the metrics at CS+UD showed discrepancies: while ECE slightly increased for the Histogram calibrator (showing a worse calibration quality), it remained the same for the Temperature calibrator. On the other hand, PCS, APCS, and MPCS decreased (signaling a worse calibration quality) for both the Histogram and Temperature calibrator. Furthermore, the decrease in the metrics' values was more pronounced for the Histogram calibrator. When visually assessing both calibrators, we found that they had a similar initial distribution (CS), but the Histogram calibrator ended up much more skewed than the Temperature calibrator at CS+UD. While the ECE metric did not capture this behavior, it was successfully summarized in the PCS, APCS, and MPCS metrics. We found the same patterns could be observed when analyzing the weighted metrics (see Table \ref{T:RESULTS-WEIGHTED}).

From the results above, we confirm that the proposed metrics can accurately measure the quality of calibration of a given calibrator when no ground truth is available. Furthermore, the metrics have shown to provide a more accurate measurement of the calibrators' quality than ECE, overcoming some of its shortcomings (e.g., providing a more holistic view of the distribution of positives along the mean predicted probability, taking into account empty bins). 

Our research shows that tracking predictions over time did not enhance the quality of calibration for any of the methods involved (Histogram calibration, Platt calibration, or Temperature calibration). Finding accurate calibration models for probability calibration, given a lack of ground truth, remains a matter of future work.

\begin{table*}[ht!]
\resizebox{\textwidth}{!}{
\begin{tabular}{|l|l|rr|rr|}
\hline
\multirow{2}{*}{\textbf{Source of truth}} & \textbf{Correlation} & \multicolumn{2}{c|}{\textbf{Pearson}} & \multicolumn{2}{c|}{\textbf{Kendall}} \\ \cline{2-6} 
 & \textbf{Calibration   data} & \multicolumn{1}{l|}{\textbf{CS}} & \multicolumn{1}{l|}{\textbf{CS+UD}} & \multicolumn{1}{l|}{\textbf{CS}} & \multicolumn{1}{l|}{\textbf{CS+UD}} \\ \hline
\multirow{4}{*}{\textbf{Golden   standard}} & \textbf{ECE vs. PCS} & \multicolumn{1}{r|}{-0.6100} & -0.5937 & \multicolumn{1}{r|}{-0.3360} & -0.3981 \\ \cline{2-6} 
 & \textbf{wECE vs. wPCS} & \multicolumn{1}{r|}{-0.5113} & -0.5070 & \multicolumn{1}{r|}{-0.3574} & -0.3525 \\ \cline{2-6} 
 & \textbf{ECE vs. APCS\textsubscript{W}} & \multicolumn{1}{r|}{-0.6100} & -0.5939 & \multicolumn{1}{r|}{-0.3316} & -0.3981 \\ \cline{2-6} 
 & \textbf{wECE vs. wAPCS\textsubscript{W}} & \multicolumn{1}{r|}{-0.5112} & -0.5070 & \multicolumn{1}{r|}{0.3574} & -0.3480 \\ \hline
\multirow{4}{*}{\textbf{Predicted   labels}} & \textbf{ECE vs. PCS} & \multicolumn{1}{r|}{-0.6100} & -0.5937 & \multicolumn{1}{r|}{-0.3360} & -0.3981 \\ \cline{2-6} 
 & \textbf{wECE vs. wPCS} & \multicolumn{1}{r|}{-0.5084} & -0.5017 & \multicolumn{1}{r|}{-0.3360} & -0.3308 \\ \cline{2-6} 
 & \textbf{ECE vs. APCS\textsubscript{W}} & \multicolumn{1}{r|}{-0.6100} & -0.5939 & \multicolumn{1}{r|}{-0.3316} & -0.3981 \\ \cline{2-6} 
 & \textbf{wECE vs. wAPCS\textsubscript{W}} & \multicolumn{1}{r|}{-0.5083} & -0.5016 & \multicolumn{1}{r|}{0.3423} & -0.3308 \\ \hline
\end{tabular}
\caption{The results were obtained for different models and probability calibration techniques. We show Pearson and Kendall correlation coefficients when comparing the ECE, APCS\textsubscript{W}, wAPCS\textsubscript{W}, PCS, and wPCS metrics. APCS\textsubscript{W}, wAPCS\textsubscript{W}, PCS, and wPCS are measured considering the ground truth (golden standard) or approximate ground truth (predicted labels). The ECE metric is always computed considering the ground truth. \textit{CS} stands for \textit{Calibration Set}, while \textit{CS+UD} abbreviates \textit{Calibration Set + Unlabeled Data}.\label{T:RESULTS-CORRELATIONS}}}
\end{table*}

\begin{table*}[ht!]
\resizebox{\textwidth}{!}{
\begin{tabular}{|l|l|rr|rr|rr|rr|rr|}
\hline
\multirow{2}{*}{\textbf{Source of truth}} &  & \multicolumn{2}{c|}{\textbf{AUC ROC (up)}} & \multicolumn{2}{c|}{\textbf{ECE (down)}} & \multicolumn{2}{c|}{\textbf{PCS (up)}} & \multicolumn{2}{c|}{\textbf{APCS (up)}} & \multicolumn{2}{c|}{\textbf{MPCS (up)}} \\ \cline{2-12} 
 & \textbf{Calibration} & \multicolumn{1}{c|}{\textbf{CS}} & \multicolumn{1}{c|}{\textbf{CS+UD}} & \multicolumn{1}{c|}{\textbf{CS}} & \multicolumn{1}{c|}{\textbf{CS+UD}} & \multicolumn{1}{c|}{\textbf{CS}} & \multicolumn{1}{c|}{\textbf{CS+UD}} & \multicolumn{1}{c|}{\textbf{CS}} & \multicolumn{1}{c|}{\textbf{CS+UD}} & \multicolumn{1}{c|}{\textbf{CS}} & \multicolumn{1}{c|}{\textbf{CS+UD}} \\ \hline
\multirow{4}{*}{\textbf{Golden standard}} & \textbf{None} & \multicolumn{1}{r|}{0.8630} & 0.8630 & \multicolumn{1}{r|}{0.1090} & 0.1090 & \multicolumn{1}{r|}{0.7636} & 0.7636 & \multicolumn{1}{r|}{0.7448} & 0.7448 & \multicolumn{1}{r|}{0.5647} & 0.5647 \\ \cline{2-12} 
 & \textbf{Histogram} & \multicolumn{1}{r|}{0.8432} & 0.8442 & \multicolumn{1}{r|}{0.1051} & 0.1084 & \multicolumn{1}{r|}{0.8050} & 0.6509 & \multicolumn{1}{r|}{0.7458} & 0.6697 & \multicolumn{1}{r|}{0.5539} & 0.4507 \\ \cline{2-12} 
 & \textbf{Platt} & \multicolumn{1}{r|}{\textbf{0.8907}} & 0.8903 & \multicolumn{1}{r|}{\textbf{0.0914}} & 0.0955 & \multicolumn{1}{r|}{0.7523} & 0.7421 & \multicolumn{1}{r|}{\textbf{0.7669}} & 0.7613 & \multicolumn{1}{r|}{\textbf{0.5904}} & 0.5820 \\ \cline{2-12} 
 & \textbf{Temperature} & \multicolumn{1}{r|}{0.8614} & 0.8609 & \multicolumn{1}{r|}{0.1090} & 0.1090 & \multicolumn{1}{r|}{\textbf{0.8073}} & 0.7548 & \multicolumn{1}{r|}{0.7651} & 0.7383 & \multicolumn{1}{r|}{0.5886} & 0.5517 \\ \hline
\multirow{4}{*}{\textbf{Predicted labels}} & \textbf{None} & \multicolumn{1}{r|}{0.8630} & 0.8630 & \multicolumn{1}{r|}{0.1090} & 0.1090 & \multicolumn{1}{r|}{0.7636} & 0.7636 & \multicolumn{1}{r|}{0.7448} & 0.7448 & \multicolumn{1}{r|}{0.5647} & 0.5647 \\ \cline{2-12} 
 & \textbf{Histogram} & \multicolumn{1}{r|}{0.8432} & 0.8442 & \multicolumn{1}{r|}{0.1051} & 0.1084 & \multicolumn{1}{r|}{0.8050} & 0.6509 & \multicolumn{1}{r|}{0.7458} & 0.6697 & \multicolumn{1}{r|}{0.5539} & 0.4507 \\ \cline{2-12} 
 & \textbf{Platt} & \multicolumn{1}{r|}{\textbf{0.8907}} & 0.8903 & \multicolumn{1}{r|}{\textbf{0.0914}} & 0.0955 & \multicolumn{1}{r|}{0.7523} & 0.7421 & \multicolumn{1}{r|}{\textbf{0.7669}} & 0.7613 & \multicolumn{1}{r|}{\textbf{0.5904}} & 0.5820 \\ \cline{2-12} 
 & \textbf{Temperature} & \multicolumn{1}{r|}{0.8614} & 0.8609 & \multicolumn{1}{r|}{0.1090} & 0.1090 & \multicolumn{1}{r|}{\textbf{0.8073}} & 0.7548 & \multicolumn{1}{r|}{0.7651} & 0.7383 & \multicolumn{1}{r|}{0.5886} & 0.5517 \\ \hline
\end{tabular}
\caption{The results were obtained for different probability calibration techniques. PCS, APCS, and MPCS are measured considering the ground truth (golden standard) or approximate ground truth (predicted labels). The AUC ROC and ECE metrics are always computed considering the ground truth. The best results are bolded. \label{T:RESULTS-NOT-WEIGHTED}}}
\end{table*}

\begin{table*}[ht!]
\resizebox{\textwidth}{!}{
\begin{tabular}{|l|l|rr|rr|rr|rr|rr|}
\hline
\multirow{2}{*}{\textbf{Source of truth}} &  & \multicolumn{2}{c|}{\textbf{AUC ROC (up)}} & \multicolumn{2}{c|}{\textbf{wECE (down)}} & \multicolumn{2}{c|}{\textbf{wPCS (up)}} & \multicolumn{2}{c|}{\textbf{wAPCS (up)}} & \multicolumn{2}{c|}{\textbf{wMPCS (up)}} \\ \cline{2-12} 
 & \textbf{Calibration} & \multicolumn{1}{c|}{\textbf{CS}} & \multicolumn{1}{c|}{\textbf{CS+UD}} & \multicolumn{1}{c|}{\textbf{CS}} & \multicolumn{1}{c|}{\textbf{CS+UD}} & \multicolumn{1}{c|}{\textbf{CS}} & \multicolumn{1}{c|}{\textbf{CS+UD}} & \multicolumn{1}{c|}{\textbf{CS}} & \multicolumn{1}{c|}{\textbf{CS+UD}} & \multicolumn{1}{c|}{\textbf{CS}} & \multicolumn{1}{c|}{\textbf{CS+UD}} \\ \hline
\multirow{4}{*}{\textbf{Golden standard}} & \textbf{None} & \multicolumn{1}{r|}{0.8630} & 0.8630 & \multicolumn{1}{r|}{0.1457} & 0.1457 & \multicolumn{1}{r|}{0.7829} & 0.7829 & \multicolumn{1}{r|}{0.7545} & 0.7545 & \multicolumn{1}{r|}{0.5801} & 0.5801 \\ \cline{2-12} 
 & \textbf{Histogram} & \multicolumn{1}{r|}{0.8432} & 0.8442 & \multicolumn{1}{r|}{0.1410} & 0.1449 & \multicolumn{1}{r|}{\textbf{0.8265}} & 0.6494 & \multicolumn{1}{r|}{0.7565} & 0.6689 & \multicolumn{1}{r|}{0.5685} & 0.4505 \\ \cline{2-12} 
 & \textbf{Platt} & \multicolumn{1}{r|}{\textbf{0.8907}} & 0.8903 & \multicolumn{1}{r|}{\textbf{0.1230}} & 0.1285 & \multicolumn{1}{r|}{0.7655} & 0.7527 & \multicolumn{1}{r|}{\textbf{0.7735}} & 0.7666 & \multicolumn{1}{r|}{0.6010} & 0.5902 \\ \cline{2-12} 
 & \textbf{Temperature} & \multicolumn{1}{r|}{0.8614} & 0.8609 & \multicolumn{1}{r|}{0.1457} & 0.1457 & \multicolumn{1}{r|}{0.8232} & 0.7773 & \multicolumn{1}{r|}{0.7730} & 0.7496 & \multicolumn{1}{r|}{\textbf{0.6014}} & 0.5697 \\ \hline
\multirow{4}{*}{\textbf{Predicted labels}} & \textbf{None} & \multicolumn{1}{r|}{0.8630} & 0.8630 & \multicolumn{1}{r|}{0.1457} & 0.1457 & \multicolumn{1}{r|}{0.7826} & 0.7826 & \multicolumn{1}{r|}{0.7543} & 0.7543 & \multicolumn{1}{r|}{0.5799} & 0.5799 \\ \cline{2-12} 
 & \textbf{Histogram} & \multicolumn{1}{r|}{0.8432} & 0.8442 & \multicolumn{1}{r|}{0.1410} & 0.1449 & \multicolumn{1}{r|}{\textbf{0.8265}} & 0.6494 & \multicolumn{1}{r|}{0.7565} & 0.6689 & \multicolumn{1}{r|}{0.5686} & 0.4505 \\ \cline{2-12} 
 & \textbf{Platt} & \multicolumn{1}{r|}{\textbf{0.8907}} & 0.8903 & \multicolumn{1}{r|}{\textbf{0.1230}} & 0.1285 & \multicolumn{1}{r|}{0.7641} & 0.7517 & \multicolumn{1}{r|}{0.7728} & 0.7662 & \multicolumn{1}{r|}{0.5998} & 0.5894 \\ \cline{2-12} 
 & \textbf{Temperature} & \multicolumn{1}{r|}{0.8614} & 0.8609 & \multicolumn{1}{r|}{0.1457} & 0.1457 & \multicolumn{1}{r|}{0.8230} & 0.7765 & \multicolumn{1}{r|}{0.7730} & 0.7492 & \multicolumn{1}{r|}{0.6013} & 0.5692 \\ \hline
\end{tabular}
\caption{The results were obtained for different probability calibration techniques. wPCS, wAPCS, and wMPCS are measured considering the ground truth (golden standard) or approximate ground truth (predicted labels). The AUC ROC and wECE metrics are always computed considering the ground truth. The best results are bolded. \label{T:RESULTS-WEIGHTED}}}
\end{table*}

\section{Conclusions and Future Work}\label{S:CONCLUSION}
This work explored active learning with multiple oracles to alleviate the manual inspection of manufactured products and the labeling of inspected products. Our active learning settings can save up to four percent of the manual inspection and data labeling load while not compromising on the quality of the outcome for a quality threshold of p=0.95. It must be noted that labeling savings depend on the machine learning model deployed, the acceptance quality levels, and the quality of the active learning machine oracles under consideration. Furthermore, multiple probability calibration techniques were compared, and several new metrics to measure the quality of a calibrator were proposed. The metrics enable measuring the calibrators' quality even when no ground truth is available. The experiments demonstrated that the proposed metrics capture relevant data otherwise summarized in the ECE metric - a popular metric to measure the quality of a probability calibration model. Nevertheless, the behavior of the proposed metrics under concept drift was not studied yet, and we consider it a matter of future research.

We envision multiple lines of investigation for future work. Regarding active learning, we are interested in enriching our current setup by adopting different strategies to decide how interesting an upcoming image is (e.g., learning distance metrics for each class or learning to predict which piece of data would enhance the classifier the most) and enhancing the calibration techniques to display the desired behavior for high-confidence thresholds. We will conduct further research on probabilities calibration to understand how the proposed metrics behave when concept drift occurs. Finally, we will explore new approximate probability calibration approaches leading to enhanced calibrators when no ground truth is available.

\bmhead{Acknowledgments}
This work was supported by the Slovenian Research Agency and the European Union's Horizon 2020 project STAR under grant agreement H2020-956573.

\section*{Declarations}
\subsection*{Conflict of interest}
The authors have no competing interests to declare relevant to this article's content.

\subsection*{Availability of data and materials}
The datasets analyzed during the current study are not publicly available for confidentiality reasons.

\subsection*{Author contributions}
Jo\v{z}e M. Ro\v{z}anec: Conceptualization, Methodology, Software Programming, Validation, Formal Analysis, Investigation, Writing (Original Draft, Review and Editing), Visualization, Supervision. Luka Bizjak: Conceptualization, Validation, Writing (Review and Editing). Elena Trajkova: Software Programming, Writing (Review and Editing). Patrik Zajec: Software Programming, Writing (Review and Editing). Jelle Keizer: Resources, Data Curation. Bla\v{z} Fortuna: Resources, Writing (Review and Editing), Supervision, Project administration, Funding acquisition. Dunja Mladeni\'{c}: Resources, Writing (Review and Editing), Supervision, Project administration, Funding acquisition.









\bibliography{main}

\begin{thebibliography}{}
\providecommand{\doi}[1]{\url{https://doi.org/#1}}
\bibcommenthead

\bibitem [\protect \citeauthoryear {%
Aggour%
\ \protect \BOthers {.}}{%
Aggour%
\ \protect \BOthers {.}}{%
{\protect \APACyear {2019}}%
}]{%
aggour2019artificial}
\APACinsertmetastar {%
aggour2019artificial}%
\begin{APACrefauthors}%
Aggour, K.S.%
, Gupta, V.K.%
, Ruscitto, D.%
, Ajdelsztajn, L.%
, Bian, X.%
, Brosnan, K.H.%
\BDBL {}others%
\end{APACrefauthors}%
\unskip\
\newblock
\APACrefYearMonthDay{2019}{}{}.
\newblock
{\BBOQ}\APACrefatitle {Artificial intelligence/machine learning in
  manufacturing and inspection: A GE perspective} {Artificial
  intelligence/machine learning in manufacturing and inspection: A ge
  perspective}.{\BBCQ}
\newblock
\APACjournalVolNumPages{MRS Bulletin}{44}{7}{545--558}.
\newblock

\newblock

\PrintBackRefs{\CurrentBib}

\bibitem [\protect \citeauthoryear {%
Aiger%
\ \BBA {} Talbot%
}{%
Aiger%
\ \BBA {} Talbot%
}{%
{\protect \APACyear {2012}}%
}]{%
aiger2012phase}
\APACinsertmetastar {%
aiger2012phase}%
\begin{APACrefauthors}%
Aiger, D.%
\BCBT {}\ \BBA {} Talbot, H.%
\end{APACrefauthors}%
\unskip\
\newblock
\APACrefYearMonthDay{2012}{}{}.
\newblock
{\BBOQ}\APACrefatitle {The phase only transform for unsupervised surface defect
  detection} {The phase only transform for unsupervised surface defect
  detection}.{\BBCQ}
\newblock
 \APACrefbtitle {Emerging Topics In Computer Vision And Its Applications}
  {Emerging topics in computer vision and its applications}\ (\BPGS\ 215--232).
\newblock
\APACaddressPublisher{}{World Scientific}.
\PrintBackRefs{\CurrentBib}

\bibitem [\protect \citeauthoryear {%
Beltr{\'a}n-Gonz{\'a}lez%
, Bustreo%
\BCBL {}\ \BBA {} Del~Bue%
}{%
Beltr{\'a}n-Gonz{\'a}lez%
\ \protect \BOthers {.}}{%
{\protect \APACyear {2020}}%
}]{%
beltran2020external}
\APACinsertmetastar {%
beltran2020external}%
\begin{APACrefauthors}%
Beltr{\'a}n-Gonz{\'a}lez, C.%
, Bustreo, M.%
\BCBL {} Del~Bue, A.%
\end{APACrefauthors}%
\unskip\
\newblock
\APACrefYearMonthDay{2020}{}{}.
\newblock
{\BBOQ}\APACrefatitle {External and internal quality inspection of aerospace
  components} {External and internal quality inspection of aerospace
  components}.{\BBCQ}
\newblock
 \APACrefbtitle {2020 IEEE 7th International Workshop on Metrology for
  AeroSpace (MetroAeroSpace)} {2020 ieee 7th international workshop on
  metrology for aerospace (metroaerospace)}\ (\BPGS\ 351--355).
\PrintBackRefs{\CurrentBib}

\bibitem [\protect \citeauthoryear {%
Beluch%
, Genewein%
, N{\"u}rnberger%
\BCBL {}\ \BBA {} K{\"o}hler%
}{%
Beluch%
\ \protect \BOthers {.}}{%
{\protect \APACyear {2018}}%
}]{%
beluch2018power}
\APACinsertmetastar {%
beluch2018power}%
\begin{APACrefauthors}%
Beluch, W.H.%
, Genewein, T.%
, N{\"u}rnberger, A.%
\BCBL {} K{\"o}hler, J.M.%
\end{APACrefauthors}%
\unskip\
\newblock
\APACrefYearMonthDay{2018}{}{}.
\newblock
{\BBOQ}\APACrefatitle {The power of ensembles for active learning in image
  classification} {The power of ensembles for active learning in image
  classification}.{\BBCQ}
\newblock
 \APACrefbtitle {Proceedings of the IEEE Conference on Computer Vision and
  Pattern Recognition} {Proceedings of the ieee conference on computer vision
  and pattern recognition}\ (\BPGS\ 9368--9377).
\PrintBackRefs{\CurrentBib}

\bibitem [\protect \citeauthoryear {%
Bradley%
}{%
Bradley%
}{%
{\protect \APACyear {1997}}%
}]{%
BRADLEY19971145}
\APACinsertmetastar {%
BRADLEY19971145}%
\begin{APACrefauthors}%
Bradley, A.P.%
\end{APACrefauthors}%
\unskip\
\newblock
\APACrefYearMonthDay{1997}{}{}.
\newblock
{\BBOQ}\APACrefatitle {The use of the area under the ROC curve in the
  evaluation of machine learning algorithms} {The use of the area under the roc
  curve in the evaluation of machine learning algorithms}.{\BBCQ}
\newblock
\APACjournalVolNumPages{Pattern Recognition}{30}{7}{1145 - 1159}.
\newblock
\begin{APACrefURL}
  {http://www.sciencedirect.com/science/article/pii/S0031320396001422}
  \end{APACrefURL}
\newblock

\newblock

\newblock
\begin{APACrefDOI} \doi{https://doi.org/10.1016/S0031-3203(96)00142-2}
  \end{APACrefDOI}
\PrintBackRefs{\CurrentBib}

\bibitem [\protect \citeauthoryear {%
Br{\"o}cker%
\ \BBA {} Smith%
}{%
Br{\"o}cker%
\ \BBA {} Smith%
}{%
{\protect \APACyear {2007}}%
}]{%
brocker2007increasing}
\APACinsertmetastar {%
brocker2007increasing}%
\begin{APACrefauthors}%
Br{\"o}cker, J.%
\BCBT {}\ \BBA {} Smith, L.A.%
\end{APACrefauthors}%
\unskip\
\newblock
\APACrefYearMonthDay{2007}{}{}.
\newblock
{\BBOQ}\APACrefatitle {Increasing the reliability of reliability diagrams}
  {Increasing the reliability of reliability diagrams}.{\BBCQ}
\newblock
\APACjournalVolNumPages{Weather and forecasting}{22}{3}{651--661}.
\newblock

\newblock

\PrintBackRefs{\CurrentBib}

\bibitem [\protect \citeauthoryear {%
Buitinck%
\ \protect \BOthers {.}}{%
Buitinck%
\ \protect \BOthers {.}}{%
{\protect \APACyear {2013}}%
}]{%
sklearn_api}
\APACinsertmetastar {%
sklearn_api}%
\begin{APACrefauthors}%
Buitinck, L.%
, Louppe, G.%
, Blondel, M.%
, Pedregosa, F.%
, Mueller, A.%
, Grisel, O.%
\BDBL {}Varoquaux, G.%
\end{APACrefauthors}%
\unskip\
\newblock
\APACrefYearMonthDay{2013}{}{}.
\newblock
{\BBOQ}\APACrefatitle {{API} design for machine learning software: experiences
  from the scikit-learn project} {{API} design for machine learning software:
  experiences from the scikit-learn project}.{\BBCQ}
\newblock
 \APACrefbtitle {ECML PKDD Workshop: Languages for Data Mining and Machine
  Learning} {Ecml pkdd workshop: Languages for data mining and machine
  learning}\ (\BPGS\ 108--122).
\PrintBackRefs{\CurrentBib}

\bibitem [\protect \citeauthoryear {%
Carvajal~Soto%
, Tavakolizadeh%
\BCBL {}\ \BBA {} Gyulai%
}{%
Carvajal~Soto%
\ \protect \BOthers {.}}{%
{\protect \APACyear {2019}}%
}]{%
carvajal2019online}
\APACinsertmetastar {%
carvajal2019online}%
\begin{APACrefauthors}%
Carvajal~Soto, J.%
, Tavakolizadeh, F.%
\BCBL {} Gyulai, D.%
\end{APACrefauthors}%
\unskip\
\newblock
\APACrefYearMonthDay{2019}{}{}.
\newblock
{\BBOQ}\APACrefatitle {An online machine learning framework for early detection
  of product failures in an Industry 4.0 context} {An online machine learning
  framework for early detection of product failures in an industry 4.0
  context}.{\BBCQ}
\newblock
\APACjournalVolNumPages{International Journal of Computer Integrated
  Manufacturing}{32}{4-5}{452--465}.
\newblock

\newblock

\PrintBackRefs{\CurrentBib}

\bibitem [\protect \citeauthoryear {%
Cheeseman%
}{%
Cheeseman%
}{%
{\protect \APACyear {1985}}%
}]{%
cheeseman1985defense}
\APACinsertmetastar {%
cheeseman1985defense}%
\begin{APACrefauthors}%
Cheeseman, P.C.%
\end{APACrefauthors}%
\unskip\
\newblock
\APACrefYearMonthDay{1985}{}{}.
\newblock
{\BBOQ}\APACrefatitle {In Defense of Probability.} {In defense of
  probability.}{\BBCQ}
\newblock
 \APACrefbtitle {IJCAI} {Ijcai}\ (\BVOL~85, \BPGS\ 1002--1009).
\PrintBackRefs{\CurrentBib}

\bibitem [\protect \citeauthoryear {%
Chouchene%
\ \protect \BOthers {.}}{%
Chouchene%
\ \protect \BOthers {.}}{%
{\protect \APACyear {2020}}%
}]{%
chouchene2020artificial}
\APACinsertmetastar {%
chouchene2020artificial}%
\begin{APACrefauthors}%
Chouchene, A.%
, Carvalho, A.%
, Lima, T.M.%
, Charrua-Santos, F.%
, Os{\'o}rio, G.J.%
\BCBL {} Barhoumi, W.%
\end{APACrefauthors}%
\unskip\
\newblock
\APACrefYearMonthDay{2020}{}{}.
\newblock
{\BBOQ}\APACrefatitle {Artificial intelligence for product quality inspection
  toward smart industries: quality control of vehicle non-conformities}
  {Artificial intelligence for product quality inspection toward smart
  industries: quality control of vehicle non-conformities}.{\BBCQ}
\newblock
 \APACrefbtitle {2020 9th international conference on industrial technology and
  management (ICITM)} {2020 9th international conference on industrial
  technology and management (icitm)}\ (\BPGS\ 127--131).
\PrintBackRefs{\CurrentBib}

\bibitem [\protect \citeauthoryear {%
I.~Cohen%
\ \BBA {} Goldszmidt%
}{%
I.~Cohen%
\ \BBA {} Goldszmidt%
}{%
{\protect \APACyear {2004}}%
}]{%
cohen2004properties}
\APACinsertmetastar {%
cohen2004properties}%
\begin{APACrefauthors}%
Cohen, I.%
\BCBT {}\ \BBA {} Goldszmidt, M.%
\end{APACrefauthors}%
\unskip\
\newblock
\APACrefYearMonthDay{2004}{}{}.
\newblock
{\BBOQ}\APACrefatitle {Properties and benefits of calibrated classifiers}
  {Properties and benefits of calibrated classifiers}.{\BBCQ}
\newblock
 \APACrefbtitle {European Conference on Principles of Data Mining and Knowledge
  Discovery} {European conference on principles of data mining and knowledge
  discovery}\ (\BPGS\ 125--136).
\PrintBackRefs{\CurrentBib}

\bibitem [\protect \citeauthoryear {%
N.~Cohen%
\ \BBA {} Hoshen%
}{%
N.~Cohen%
\ \BBA {} Hoshen%
}{%
{\protect \APACyear {2020}}%
}]{%
cohen2020sub}
\APACinsertmetastar {%
cohen2020sub}%
\begin{APACrefauthors}%
Cohen, N.%
\BCBT {}\ \BBA {} Hoshen, Y.%
\end{APACrefauthors}%
\unskip\
\newblock
\APACrefYearMonthDay{2020}{}{}.
\newblock
{\BBOQ}\APACrefatitle {Sub-image anomaly detection with deep pyramid
  correspondences} {Sub-image anomaly detection with deep pyramid
  correspondences}.{\BBCQ}
\newblock
\APACjournalVolNumPages{arXiv preprint arXiv:2005.02357}{}{}{}.
\newblock

\newblock

\PrintBackRefs{\CurrentBib}

\bibitem [\protect \citeauthoryear {%
Cohn%
, Atlas%
\BCBL {}\ \BBA {} Ladner%
}{%
Cohn%
\ \protect \BOthers {.}}{%
{\protect \APACyear {1994}}%
}]{%
cohn1994improving}
\APACinsertmetastar {%
cohn1994improving}%
\begin{APACrefauthors}%
Cohn, D.%
, Atlas, L.%
\BCBL {} Ladner, R.%
\end{APACrefauthors}%
\unskip\
\newblock
\APACrefYearMonthDay{1994}{}{}.
\newblock
{\BBOQ}\APACrefatitle {Improving generalization with active learning}
  {Improving generalization with active learning}.{\BBCQ}
\newblock
\APACjournalVolNumPages{Machine learning}{15}{2}{201--221}.
\newblock

\newblock

\PrintBackRefs{\CurrentBib}

\bibitem [\protect \citeauthoryear {%
Cordier%
, Das%
\BCBL {}\ \BBA {} Gutierrez%
}{%
Cordier%
\ \protect \BOthers {.}}{%
{\protect \APACyear {2021}}%
}]{%
cordier2021active}
\APACinsertmetastar {%
cordier2021active}%
\begin{APACrefauthors}%
Cordier, A.%
, Das, D.%
\BCBL {} Gutierrez, P.%
\end{APACrefauthors}%
\unskip\
\newblock
\APACrefYearMonthDay{2021}{}{}.
\newblock
{\BBOQ}\APACrefatitle {Active learning using weakly supervised signals for
  quality inspection} {Active learning using weakly supervised signals for
  quality inspection}.{\BBCQ}
\newblock
\APACjournalVolNumPages{arXiv preprint arXiv:2104.02973}{}{}{}.
\newblock

\newblock

\PrintBackRefs{\CurrentBib}

\bibitem [\protect \citeauthoryear {%
Cullinane%
, Bosak%
, Flood%
\BCBL {}\ \BBA {} Demerouti%
}{%
Cullinane%
\ \protect \BOthers {.}}{%
{\protect \APACyear {2013}}%
}]{%
cullinane2013job}
\APACinsertmetastar {%
cullinane2013job}%
\begin{APACrefauthors}%
Cullinane, S\BHBI J.%
, Bosak, J.%
, Flood, P.C.%
\BCBL {} Demerouti, E.%
\end{APACrefauthors}%
\unskip\
\newblock
\APACrefYearMonthDay{2013}{}{}.
\newblock
{\BBOQ}\APACrefatitle {Job design under lean manufacturing and its impact on
  employee outcomes} {Job design under lean manufacturing and its impact on
  employee outcomes}.{\BBCQ}
\newblock
\APACjournalVolNumPages{Organizational Psychology Review}{3}{1}{41--61}.
\newblock

\newblock

\PrintBackRefs{\CurrentBib}

\bibitem [\protect \citeauthoryear {%
Dai%
, Mujeeb%
, Erdt%
\BCBL {}\ \BBA {} Sourin%
}{%
Dai%
\ \protect \BOthers {.}}{%
{\protect \APACyear {2018}}%
}]{%
dai2018towards}
\APACinsertmetastar {%
dai2018towards}%
\begin{APACrefauthors}%
Dai, W.%
, Mujeeb, A.%
, Erdt, M.%
\BCBL {} Sourin, A.%
\end{APACrefauthors}%
\unskip\
\newblock
\APACrefYearMonthDay{2018}{}{}.
\newblock
{\BBOQ}\APACrefatitle {Towards automatic optical inspection of soldering
  defects} {Towards automatic optical inspection of soldering defects}.{\BBCQ}
\newblock
 \APACrefbtitle {2018 International Conference on Cyberworlds (CW)} {2018
  international conference on cyberworlds (cw)}\ (\BPGS\ 375--382).
\PrintBackRefs{\CurrentBib}

\bibitem [\protect \citeauthoryear {%
Duan%
, Wang%
, Liu%
\BCBL {}\ \BBA {} Chen%
}{%
Duan%
\ \protect \BOthers {.}}{%
{\protect \APACyear {2012}}%
}]{%
duan2012machine}
\APACinsertmetastar {%
duan2012machine}%
\begin{APACrefauthors}%
Duan, G.%
, Wang, H.%
, Liu, Z.%
\BCBL {} Chen, Y\BHBI W.%
\end{APACrefauthors}%
\unskip\
\newblock
\APACrefYearMonthDay{2012}{}{}.
\newblock
{\BBOQ}\APACrefatitle {A machine learning-based framework for automatic visual
  inspection of microdrill bits in PCB production} {A machine learning-based
  framework for automatic visual inspection of microdrill bits in pcb
  production}.{\BBCQ}
\newblock
\APACjournalVolNumPages{IEEE Transactions on Systems, Man, and Cybernetics,
  Part C (Applications and Reviews)}{42}{6}{1679--1689}.
\newblock

\newblock

\PrintBackRefs{\CurrentBib}

\bibitem [\protect \citeauthoryear {%
Guo%
, Pleiss%
, Sun%
\BCBL {}\ \BBA {} Weinberger%
}{%
Guo%
\ \protect \BOthers {.}}{%
{\protect \APACyear {2017}}%
}]{%
guo2017calibration}
\APACinsertmetastar {%
guo2017calibration}%
\begin{APACrefauthors}%
Guo, C.%
, Pleiss, G.%
, Sun, Y.%
\BCBL {} Weinberger, K.Q.%
\end{APACrefauthors}%
\unskip\
\newblock
\APACrefYearMonthDay{2017}{}{}.
\newblock
{\BBOQ}\APACrefatitle {On calibration of modern neural networks} {On
  calibration of modern neural networks}.{\BBCQ}
\newblock
 \APACrefbtitle {International conference on machine learning} {International
  conference on machine learning}\ (\BPGS\ 1321--1330).
\PrintBackRefs{\CurrentBib}

\bibitem [\protect \citeauthoryear {%
Gupta%
\ \BBA {} Ramdas%
}{%
Gupta%
\ \BBA {} Ramdas%
}{%
{\protect \APACyear {2021}}%
}]{%
gupta2021distribution}
\APACinsertmetastar {%
gupta2021distribution}%
\begin{APACrefauthors}%
Gupta, C.%
\BCBT {}\ \BBA {} Ramdas, A.%
\end{APACrefauthors}%
\unskip\
\newblock
\APACrefYearMonthDay{2021}{}{}.
\newblock
{\BBOQ}\APACrefatitle {Distribution-free calibration guarantees for histogram
  binning without sample splitting} {Distribution-free calibration guarantees
  for histogram binning without sample splitting}.{\BBCQ}
\newblock
 \APACrefbtitle {International Conference on Machine Learning} {International
  conference on machine learning}\ (\BPGS\ 3942--3952).
\PrintBackRefs{\CurrentBib}

\bibitem [\protect \citeauthoryear {%
He%
, Zhang%
, Ren%
\BCBL {}\ \BBA {} Sun%
}{%
He%
\ \protect \BOthers {.}}{%
{\protect \APACyear {2016}}%
}]{%
he2016deep}
\APACinsertmetastar {%
he2016deep}%
\begin{APACrefauthors}%
He, K.%
, Zhang, X.%
, Ren, S.%
\BCBL {} Sun, J.%
\end{APACrefauthors}%
\unskip\
\newblock
\APACrefYearMonthDay{2016}{}{}.
\newblock
{\BBOQ}\APACrefatitle {Deep residual learning for image recognition} {Deep
  residual learning for image recognition}.{\BBCQ}
\newblock
 \APACrefbtitle {Proceedings of the IEEE conference on computer vision and
  pattern recognition} {Proceedings of the ieee conference on computer vision
  and pattern recognition}\ (\BPGS\ 770--778).
\PrintBackRefs{\CurrentBib}

\bibitem [\protect \citeauthoryear {%
Hua%
, Xiong%
, Lowey%
, Suh%
\BCBL {}\ \BBA {} Dougherty%
}{%
Hua%
\ \protect \BOthers {.}}{%
{\protect \APACyear {2005}}%
}]{%
hua2005optimal}
\APACinsertmetastar {%
hua2005optimal}%
\begin{APACrefauthors}%
Hua, J.%
, Xiong, Z.%
, Lowey, J.%
, Suh, E.%
\BCBL {} Dougherty, E.R.%
\end{APACrefauthors}%
\unskip\
\newblock
\APACrefYearMonthDay{2005}{}{}.
\newblock
{\BBOQ}\APACrefatitle {Optimal number of features as a function of sample size
  for various classification rules} {Optimal number of features as a function
  of sample size for various classification rules}.{\BBCQ}
\newblock
\APACjournalVolNumPages{Bioinformatics}{21}{8}{1509--1515}.
\newblock

\newblock

\PrintBackRefs{\CurrentBib}

\bibitem [\protect \citeauthoryear {%
Jezek%
, Jonak%
, Burget%
, Dvorak%
\BCBL {}\ \BBA {} Skotak%
}{%
Jezek%
\ \protect \BOthers {.}}{%
{\protect \APACyear {2021}}%
}]{%
jezek2021deep}
\APACinsertmetastar {%
jezek2021deep}%
\begin{APACrefauthors}%
Jezek, S.%
, Jonak, M.%
, Burget, R.%
, Dvorak, P.%
\BCBL {} Skotak, M.%
\end{APACrefauthors}%
\unskip\
\newblock
\APACrefYearMonthDay{2021}{}{}.
\newblock
{\BBOQ}\APACrefatitle {Deep learning-based defect detection of metal parts:
  evaluating current methods in complex conditions} {Deep learning-based defect
  detection of metal parts: evaluating current methods in complex
  conditions}.{\BBCQ}
\newblock
 \APACrefbtitle {2021 13th International Congress on Ultra Modern
  Telecommunications and Control Systems and Workshops (ICUMT)} {2021 13th
  international congress on ultra modern telecommunications and control systems
  and workshops (icumt)}\ (\BPGS\ 66--71).
\PrintBackRefs{\CurrentBib}

\bibitem [\protect \citeauthoryear {%
Jian%
, Gao%
\BCBL {}\ \BBA {} Ao%
}{%
Jian%
\ \protect \BOthers {.}}{%
{\protect \APACyear {2017}}%
}]{%
jian2017automatic}
\APACinsertmetastar {%
jian2017automatic}%
\begin{APACrefauthors}%
Jian, C.%
, Gao, J.%
\BCBL {} Ao, Y.%
\end{APACrefauthors}%
\unskip\
\newblock
\APACrefYearMonthDay{2017}{}{}.
\newblock
{\BBOQ}\APACrefatitle {Automatic surface defect detection for mobile phone
  screen glass based on machine vision} {Automatic surface defect detection for
  mobile phone screen glass based on machine vision}.{\BBCQ}
\newblock
\APACjournalVolNumPages{Applied Soft Computing}{52}{}{348--358}.
\newblock

\newblock

\PrintBackRefs{\CurrentBib}

\bibitem [\protect \citeauthoryear {%
Jiang%
\ \BBA {} Wong%
}{%
Jiang%
\ \BBA {} Wong%
}{%
{\protect \APACyear {2018}}%
}]{%
jiang2018fundamentals}
\APACinsertmetastar {%
jiang2018fundamentals}%
\begin{APACrefauthors}%
Jiang, J.%
\BCBT {}\ \BBA {} Wong, W.%
\end{APACrefauthors}%
\unskip\
\newblock
\APACrefYearMonthDay{2018}{}{}.
\newblock
{\BBOQ}\APACrefatitle {Fundamentals of common computer vision techniques for
  textile quality control} {Fundamentals of common computer vision techniques
  for textile quality control}.{\BBCQ}
\newblock
 \APACrefbtitle {Applications of Computer Vision in Fashion and Textiles}
  {Applications of computer vision in fashion and textiles}\ (\BPGS\ 3--15).
\newblock
\APACaddressPublisher{}{Elsevier}.
\PrintBackRefs{\CurrentBib}

\bibitem [\protect \citeauthoryear {%
Kang%
\ \BBA {} Liu%
}{%
Kang%
\ \BBA {} Liu%
}{%
{\protect \APACyear {2005}}%
}]{%
kang2005surface}
\APACinsertmetastar {%
kang2005surface}%
\begin{APACrefauthors}%
Kang, G\BHBI W.%
\BCBT {}\ \BBA {} Liu, H\BHBI B.%
\end{APACrefauthors}%
\unskip\
\newblock
\APACrefYearMonthDay{2005}{}{}.
\newblock
{\BBOQ}\APACrefatitle {Surface defects inspection of cold rolled strips based
  on neural network} {Surface defects inspection of cold rolled strips based on
  neural network}.{\BBCQ}
\newblock
 \APACrefbtitle {2005 International Conference on Machine Learning and
  Cybernetics} {2005 international conference on machine learning and
  cybernetics}\ (\BVOL~8, \BPGS\ 5034--5037).
\PrintBackRefs{\CurrentBib}

\bibitem [\protect \citeauthoryear {%
Kendall%
}{%
Kendall%
}{%
{\protect \APACyear {1938}}%
}]{%
kendall1938new}
\APACinsertmetastar {%
kendall1938new}%
\begin{APACrefauthors}%
Kendall, M.G.%
\end{APACrefauthors}%
\unskip\
\newblock
\APACrefYearMonthDay{1938}{}{}.
\newblock
{\BBOQ}\APACrefatitle {A new measure of rank correlation} {A new measure of
  rank correlation}.{\BBCQ}
\newblock
\APACjournalVolNumPages{Biometrika}{30}{1/2}{81--93}.
\newblock

\newblock

\PrintBackRefs{\CurrentBib}

\bibitem [\protect \citeauthoryear {%
Kuhn%
, Johnson%
\BCBL {}\ \protect \BOthers {.}}{%
Kuhn%
\ \protect \BOthers {.}}{%
{\protect \APACyear {2013}}%
}]{%
kuhn2013applied}
\APACinsertmetastar {%
kuhn2013applied}%
\begin{APACrefauthors}%
Kuhn, M.%
, Johnson, K.%
\BCBL {}\ \BOthersPeriod {.}\end{APACrefauthors}%
\unskip\
\newblock
\APACrefYear{2013}.
\newblock
\APACrefbtitle {Applied predictive modeling} {Applied predictive modeling}\
  (\BVOL~26).
\newblock
\APACaddressPublisher{}{Springer}.
\PrintBackRefs{\CurrentBib}

\bibitem [\protect \citeauthoryear {%
Kujawi{\'n}ska%
, Vogt%
\BCBL {}\ \BBA {} Hamrol%
}{%
Kujawi{\'n}ska%
\ \protect \BOthers {.}}{%
{\protect \APACyear {2016}}%
}]{%
kujawinska2016role}
\APACinsertmetastar {%
kujawinska2016role}%
\begin{APACrefauthors}%
Kujawi{\'n}ska, A.%
, Vogt, K.%
\BCBL {} Hamrol, A.%
\end{APACrefauthors}%
\unskip\
\newblock
\APACrefYearMonthDay{2016}{}{}.
\newblock
{\BBOQ}\APACrefatitle {The role of human motivation in quality inspection of
  production processes} {The role of human motivation in quality inspection of
  production processes}.{\BBCQ}
\newblock
 \APACrefbtitle {Advances in Ergonomics of Manufacturing: Managing the
  Enterprise of the Future} {Advances in ergonomics of manufacturing: Managing
  the enterprise of the future}\ (\BPGS\ 569--579).
\newblock
\APACaddressPublisher{}{Springer}.
\PrintBackRefs{\CurrentBib}

\bibitem [\protect \citeauthoryear {%
Kumar%
, Liang%
\BCBL {}\ \BBA {} Ma%
}{%
Kumar%
\ \protect \BOthers {.}}{%
{\protect \APACyear {2019}}%
}]{%
kumar2019verified}
\APACinsertmetastar {%
kumar2019verified}%
\begin{APACrefauthors}%
Kumar, A.%
, Liang, P.S.%
\BCBL {} Ma, T.%
\end{APACrefauthors}%
\unskip\
\newblock
\APACrefYearMonthDay{2019}{}{}.
\newblock
{\BBOQ}\APACrefatitle {Verified uncertainty calibration} {Verified uncertainty
  calibration}.{\BBCQ}
\newblock
\APACjournalVolNumPages{Advances in Neural Information Processing
  Systems}{32}{}{}.
\newblock

\newblock

\PrintBackRefs{\CurrentBib}

\bibitem [\protect \citeauthoryear {%
Kurniati%
, Yeh%
\BCBL {}\ \BBA {} Lin%
}{%
Kurniati%
\ \protect \BOthers {.}}{%
{\protect \APACyear {2015}}%
}]{%
kurniati2015quality}
\APACinsertmetastar {%
kurniati2015quality}%
\begin{APACrefauthors}%
Kurniati, N.%
, Yeh, R\BHBI H.%
\BCBL {} Lin, J\BHBI J.%
\end{APACrefauthors}%
\unskip\
\newblock
\APACrefYearMonthDay{2015}{}{}.
\newblock
{\BBOQ}\APACrefatitle {Quality inspection and maintenance: the framework of
  interaction} {Quality inspection and maintenance: the framework of
  interaction}.{\BBCQ}
\newblock
\APACjournalVolNumPages{Procedia manufacturing}{4}{}{244--251}.
\newblock

\newblock

\PrintBackRefs{\CurrentBib}

\bibitem [\protect \citeauthoryear {%
Küppers%
, Kronenberger%
, Shantia%
\BCBL {}\ \BBA {} Haselhoff%
}{%
Küppers%
\ \protect \BOthers {.}}{%
{\protect \APACyear {2020}}%
}]{%
Kueppers_2020_CVPR_Workshops}
\APACinsertmetastar {%
Kueppers_2020_CVPR_Workshops}%
\begin{APACrefauthors}%
Küppers, F.%
, Kronenberger, J.%
, Shantia, A.%
\BCBL {} Haselhoff, A.%
\end{APACrefauthors}%
\unskip\
\newblock
\APACrefYearMonthDay{2020}{June}{}.
\newblock
{\BBOQ}\APACrefatitle {Multivariate Confidence Calibration for Object
  Detection} {Multivariate confidence calibration for object detection}.{\BBCQ}
\newblock
 \APACrefbtitle {The IEEE/CVF Conference on Computer Vision and Pattern
  Recognition (CVPR) Workshops.} {The ieee/cvf conference on computer vision
  and pattern recognition (cvpr) workshops.}
\PrintBackRefs{\CurrentBib}

\bibitem [\protect \citeauthoryear {%
Leathart%
, Frank%
, Holmes%
\BCBL {}\ \BBA {} Pfahringer%
}{%
Leathart%
\ \protect \BOthers {.}}{%
{\protect \APACyear {2017}}%
}]{%
leathart2017probability}
\APACinsertmetastar {%
leathart2017probability}%
\begin{APACrefauthors}%
Leathart, T.%
, Frank, E.%
, Holmes, G.%
\BCBL {} Pfahringer, B.%
\end{APACrefauthors}%
\unskip\
\newblock
\APACrefYearMonthDay{2017}{}{}.
\newblock
{\BBOQ}\APACrefatitle {Probability calibration trees} {Probability calibration
  trees}.{\BBCQ}
\newblock
 \APACrefbtitle {Asian Conference on Machine Learning} {Asian conference on
  machine learning}\ (\BPGS\ 145--160).
\PrintBackRefs{\CurrentBib}

\bibitem [\protect \citeauthoryear {%
Lewis%
\ \BBA {} Catlett%
}{%
Lewis%
\ \BBA {} Catlett%
}{%
{\protect \APACyear {1994}}%
}]{%
lewis1994heterogeneous}
\APACinsertmetastar {%
lewis1994heterogeneous}%
\begin{APACrefauthors}%
Lewis, D.D.%
\BCBT {}\ \BBA {} Catlett, J.%
\end{APACrefauthors}%
\unskip\
\newblock
\APACrefYearMonthDay{1994}{}{}.
\newblock
{\BBOQ}\APACrefatitle {Heterogeneous uncertainty sampling for supervised
  learning} {Heterogeneous uncertainty sampling for supervised
  learning}.{\BBCQ}
\newblock
 \APACrefbtitle {Machine learning proceedings 1994} {Machine learning
  proceedings 1994}\ (\BPGS\ 148--156).
\newblock
\APACaddressPublisher{}{Elsevier}.
\PrintBackRefs{\CurrentBib}

\bibitem [\protect \citeauthoryear {%
Li%
, Sohn%
, Yoon%
\BCBL {}\ \BBA {} Pfister%
}{%
Li%
\ \protect \BOthers {.}}{%
{\protect \APACyear {2021}}%
}]{%
li2021cutpaste}
\APACinsertmetastar {%
li2021cutpaste}%
\begin{APACrefauthors}%
Li, C\BHBI L.%
, Sohn, K.%
, Yoon, J.%
\BCBL {} Pfister, T.%
\end{APACrefauthors}%
\unskip\
\newblock
\APACrefYearMonthDay{2021}{}{}.
\newblock
{\BBOQ}\APACrefatitle {Cutpaste: Self-supervised learning for anomaly detection
  and localization} {Cutpaste: Self-supervised learning for anomaly detection
  and localization}.{\BBCQ}
\newblock
 \APACrefbtitle {Proceedings of the IEEE/CVF Conference on Computer Vision and
  Pattern Recognition} {Proceedings of the ieee/cvf conference on computer
  vision and pattern recognition}\ (\BPGS\ 9664--9674).
\PrintBackRefs{\CurrentBib}

\bibitem [\protect \citeauthoryear {%
Meng%
\ \protect \BOthers {.}}{%
Meng%
\ \protect \BOthers {.}}{%
{\protect \APACyear {2020}}%
}]{%
meng2020machine}
\APACinsertmetastar {%
meng2020machine}%
\begin{APACrefauthors}%
Meng, L.%
, McWilliams, B.%
, Jarosinski, W.%
, Park, H\BHBI Y.%
, Jung, Y\BHBI G.%
, Lee, J.%
\BCBL {} Zhang, J.%
\end{APACrefauthors}%
\unskip\
\newblock
\APACrefYearMonthDay{2020}{}{}.
\newblock
{\BBOQ}\APACrefatitle {Machine learning in additive manufacturing: A review}
  {Machine learning in additive manufacturing: A review}.{\BBCQ}
\newblock
\APACjournalVolNumPages{Jom}{72}{6}{2363--2377}.
\newblock

\newblock

\PrintBackRefs{\CurrentBib}

\bibitem [\protect \citeauthoryear {%
Mujeeb%
, Dai%
, Erdt%
\BCBL {}\ \BBA {} Sourin%
}{%
Mujeeb%
\ \protect \BOthers {.}}{%
{\protect \APACyear {2018}}%
}]{%
mujeeb2018unsupervised}
\APACinsertmetastar {%
mujeeb2018unsupervised}%
\begin{APACrefauthors}%
Mujeeb, A.%
, Dai, W.%
, Erdt, M.%
\BCBL {} Sourin, A.%
\end{APACrefauthors}%
\unskip\
\newblock
\APACrefYearMonthDay{2018}{}{}.
\newblock
{\BBOQ}\APACrefatitle {Unsupervised surface defect detection using deep
  autoencoders and data augmentation} {Unsupervised surface defect detection
  using deep autoencoders and data augmentation}.{\BBCQ}
\newblock
 \APACrefbtitle {2018 International Conference on Cyberworlds (CW)} {2018
  international conference on cyberworlds (cw)}\ (\BPGS\ 391--398).
\PrintBackRefs{\CurrentBib}

\bibitem [\protect \citeauthoryear {%
Newman%
\ \BBA {} Jain%
}{%
Newman%
\ \BBA {} Jain%
}{%
{\protect \APACyear {1995}}%
}]{%
newman1995survey}
\APACinsertmetastar {%
newman1995survey}%
\begin{APACrefauthors}%
Newman, T.S.%
\BCBT {}\ \BBA {} Jain, A.K.%
\end{APACrefauthors}%
\unskip\
\newblock
\APACrefYearMonthDay{1995}{}{}.
\newblock
{\BBOQ}\APACrefatitle {A survey of automated visual inspection} {A survey of
  automated visual inspection}.{\BBCQ}
\newblock
\APACjournalVolNumPages{Computer vision and image
  understanding}{61}{2}{231--262}.
\newblock

\newblock

\PrintBackRefs{\CurrentBib}

\bibitem [\protect \citeauthoryear {%
Nixon%
, Dusenberry%
, Zhang%
, Jerfel%
\BCBL {}\ \BBA {} Tran%
}{%
Nixon%
\ \protect \BOthers {.}}{%
{\protect \APACyear {2019}}%
}]{%
nixon2019measuring}
\APACinsertmetastar {%
nixon2019measuring}%
\begin{APACrefauthors}%
Nixon, J.%
, Dusenberry, M.W.%
, Zhang, L.%
, Jerfel, G.%
\BCBL {} Tran, D.%
\end{APACrefauthors}%
\unskip\
\newblock
\APACrefYearMonthDay{2019}{}{}.
\newblock
{\BBOQ}\APACrefatitle {Measuring Calibration in Deep Learning.} {Measuring
  calibration in deep learning.}{\BBCQ}
\newblock
 \APACrefbtitle {CVPR Workshops} {Cvpr workshops}\ (\BVOL~2).
\PrintBackRefs{\CurrentBib}

\bibitem [\protect \citeauthoryear {%
Ovadia%
\ \protect \BOthers {.}}{%
Ovadia%
\ \protect \BOthers {.}}{%
{\protect \APACyear {2019}}%
}]{%
ovadia2019can}
\APACinsertmetastar {%
ovadia2019can}%
\begin{APACrefauthors}%
Ovadia, Y.%
, Fertig, E.%
, Ren, J.%
, Nado, Z.%
, Sculley, D.%
, Nowozin, S.%
\BDBL {}Snoek, J.%
\end{APACrefauthors}%
\unskip\
\newblock
\APACrefYearMonthDay{2019}{}{}.
\newblock
{\BBOQ}\APACrefatitle {Can you trust your model's uncertainty? evaluating
  predictive uncertainty under dataset shift} {Can you trust your model's
  uncertainty? evaluating predictive uncertainty under dataset shift}.{\BBCQ}
\newblock
\APACjournalVolNumPages{Advances in neural information processing
  systems}{32}{}{}.
\newblock

\newblock

\PrintBackRefs{\CurrentBib}

\bibitem [\protect \citeauthoryear {%
Park%
, Kwon%
, Park%
\BCBL {}\ \BBA {} Kang%
}{%
Park%
\ \protect \BOthers {.}}{%
{\protect \APACyear {2016}}%
}]{%
park2016machine}
\APACinsertmetastar {%
park2016machine}%
\begin{APACrefauthors}%
Park, J\BHBI K.%
, Kwon, B\BHBI K.%
, Park, J\BHBI H.%
\BCBL {} Kang, D\BHBI J.%
\end{APACrefauthors}%
\unskip\
\newblock
\APACrefYearMonthDay{2016}{}{}.
\newblock
{\BBOQ}\APACrefatitle {Machine learning-based imaging system for surface defect
  inspection} {Machine learning-based imaging system for surface defect
  inspection}.{\BBCQ}
\newblock
\APACjournalVolNumPages{International Journal of Precision Engineering and
  Manufacturing-Green Technology}{3}{3}{303--310}.
\newblock

\newblock

\PrintBackRefs{\CurrentBib}

\bibitem [\protect \citeauthoryear {%
Peyr{\'e}%
, Cuturi%
\BCBL {}\ \protect \BOthers {.}}{%
Peyr{\'e}%
\ \protect \BOthers {.}}{%
{\protect \APACyear {2019}}%
}]{%
peyre2019computational}
\APACinsertmetastar {%
peyre2019computational}%
\begin{APACrefauthors}%
Peyr{\'e}, G.%
, Cuturi, M.%
\BCBL {}\ \BOthersPeriod {.}\end{APACrefauthors}%
\unskip\
\newblock
\APACrefYearMonthDay{2019}{}{}.
\newblock
{\BBOQ}\APACrefatitle {Computational optimal transport: With applications to
  data science} {Computational optimal transport: With applications to data
  science}.{\BBCQ}
\newblock
\APACjournalVolNumPages{Foundations and Trends{\textregistered} in Machine
  Learning}{11}{5-6}{355--607}.
\newblock

\newblock

\PrintBackRefs{\CurrentBib}

\bibitem [\protect \citeauthoryear {%
J.~Platt%
\ \protect \BOthers {.}}{%
J.~Platt%
\ \protect \BOthers {.}}{%
{\protect \APACyear {1999}}%
}]{%
platt1999probabilistic}
\APACinsertmetastar {%
platt1999probabilistic}%
\begin{APACrefauthors}%
Platt, J.%
\BCBT {}\ \BOthersPeriod {.}
\end{APACrefauthors}%
\unskip\
\newblock
\APACrefYearMonthDay{1999}{}{}.
\newblock
{\BBOQ}\APACrefatitle {Probabilistic outputs for support vector machines and
  comparisons to regularized likelihood methods} {Probabilistic outputs for
  support vector machines and comparisons to regularized likelihood
  methods}.{\BBCQ}
\newblock
\APACjournalVolNumPages{Advances in large margin classifiers}{10}{3}{61--74}.
\newblock

\newblock

\PrintBackRefs{\CurrentBib}

\bibitem [\protect \citeauthoryear {%
J.C.~Platt%
}{%
J.C.~Platt%
}{%
{\protect \APACyear {2000}}%
}]{%
platt20005}
\APACinsertmetastar {%
platt20005}%
\begin{APACrefauthors}%
Platt, J.C.%
\end{APACrefauthors}%
\unskip\
\newblock
\APACrefYearMonthDay{2000}{}{}.
\newblock
{\BBOQ}\APACrefatitle {5 Probabilities for SV Machines} {5 probabilities for sv
  machines}.{\BBCQ}
\newblock
\APACjournalVolNumPages{Advances in Large Margin Classifiers}{}{}{61}.
\newblock

\newblock

\PrintBackRefs{\CurrentBib}

\bibitem [\protect \citeauthoryear {%
Posocco%
\ \BBA {} Bonnefoy%
}{%
Posocco%
\ \BBA {} Bonnefoy%
}{%
{\protect \APACyear {2021}}%
}]{%
posocco2021estimating}
\APACinsertmetastar {%
posocco2021estimating}%
\begin{APACrefauthors}%
Posocco, N.%
\BCBT {}\ \BBA {} Bonnefoy, A.%
\end{APACrefauthors}%
\unskip\
\newblock
\APACrefYearMonthDay{2021}{}{}.
\newblock
{\BBOQ}\APACrefatitle {Estimating Expected Calibration Errors} {Estimating
  expected calibration errors}.{\BBCQ}
\newblock
 \APACrefbtitle {International Conference on Artificial Neural Networks}
  {International conference on artificial neural networks}\ (\BPGS\ 139--150).
\PrintBackRefs{\CurrentBib}

\bibitem [\protect \citeauthoryear {%
Rai%
, Tiwari%
, Ivanov%
\BCBL {}\ \BBA {} Dolgui%
}{%
Rai%
\ \protect \BOthers {.}}{%
{\protect \APACyear {2021}}%
}]{%
rai2021machine}
\APACinsertmetastar {%
rai2021machine}%
\begin{APACrefauthors}%
Rai, R.%
, Tiwari, M.K.%
, Ivanov, D.%
\BCBL {} Dolgui, A.%
\end{APACrefauthors}%
\unskip\
\newblock
\APACrefYearMonthDay{2021}{}{}.
\newblock
\APACrefbtitle {Machine learning in manufacturing and industry 4.0
  applications} {Machine learning in manufacturing and industry 4.0
  applications}\ (\BVOL~59)\ (\BNUM~16).
\newblock
\APACaddressPublisher{}{Taylor \& Francis}.
\PrintBackRefs{\CurrentBib}

\bibitem [\protect \citeauthoryear {%
P.~Ren%
\ \protect \BOthers {.}}{%
P.~Ren%
\ \protect \BOthers {.}}{%
{\protect \APACyear {2020}}%
}]{%
ren2020survey}
\APACinsertmetastar {%
ren2020survey}%
\begin{APACrefauthors}%
Ren, P.%
, Xiao, Y.%
, Chang, X.%
, Huang, P\BHBI Y.%
, Li, Z.%
, Chen, X.%
\BCBL {} Wang, X.%
\end{APACrefauthors}%
\unskip\
\newblock
\APACrefYearMonthDay{2020}{}{}.
\newblock
{\BBOQ}\APACrefatitle {A survey of deep active learning} {A survey of deep
  active learning}.{\BBCQ}
\newblock
\APACjournalVolNumPages{arXiv preprint arXiv:2009.00236}{}{}{}.
\newblock

\newblock

\PrintBackRefs{\CurrentBib}

\bibitem [\protect \citeauthoryear {%
R.~Ren%
, Hung%
\BCBL {}\ \BBA {} Tan%
}{%
R.~Ren%
\ \protect \BOthers {.}}{%
{\protect \APACyear {2017}}%
}]{%
ren2017generic}
\APACinsertmetastar {%
ren2017generic}%
\begin{APACrefauthors}%
Ren, R.%
, Hung, T.%
\BCBL {} Tan, K.C.%
\end{APACrefauthors}%
\unskip\
\newblock
\APACrefYearMonthDay{2017}{}{}.
\newblock
{\BBOQ}\APACrefatitle {A generic deep-learning-based approach for automated
  surface inspection} {A generic deep-learning-based approach for automated
  surface inspection}.{\BBCQ}
\newblock
\APACjournalVolNumPages{IEEE transactions on cybernetics}{48}{3}{929--940}.
\newblock

\newblock

\PrintBackRefs{\CurrentBib}

\bibitem [\protect \citeauthoryear {%
Rippel%
, Haumering%
, Brauers%
\BCBL {}\ \BBA {} Merhof%
}{%
Rippel%
\ \protect \BOthers {.}}{%
{\protect \APACyear {2021}}%
}]{%
rippel2021anomaly}
\APACinsertmetastar {%
rippel2021anomaly}%
\begin{APACrefauthors}%
Rippel, O.%
, Haumering, P.%
, Brauers, J.%
\BCBL {} Merhof, D.%
\end{APACrefauthors}%
\unskip\
\newblock
\APACrefYearMonthDay{2021}{}{}.
\newblock
{\BBOQ}\APACrefatitle {Anomaly detection for the automated visual inspection of
  pet preform closures} {Anomaly detection for the automated visual inspection
  of pet preform closures}.{\BBCQ}
\newblock
 \APACrefbtitle {2021 26th IEEE International Conference on Emerging
  Technologies and Factory Automation (ETFA)} {2021 26th ieee international
  conference on emerging technologies and factory automation (etfa)}\ (\BPGS\
  1--7).
\PrintBackRefs{\CurrentBib}

\bibitem [\protect \citeauthoryear {%
Ro{\v{z}}anec%
, Novalija%
\BCBL {}\ \protect \BOthers {.}}{%
Ro{\v{z}}anec%
, Novalija%
\BCBL {}\ \protect \BOthers {.}}{%
{\protect \APACyear {2022}}%
}]{%
rovzanec2022human}
\APACinsertmetastar {%
rovzanec2022human}%
\begin{APACrefauthors}%
Ro{\v{z}}anec, J.M.%
, Novalija, I.%
, Zajec, P.%
, Kenda, K.%
, Tavakoli, H.%
, Suh, S.%
\BDBL {}others%
\end{APACrefauthors}%
\unskip\
\newblock
\APACrefYearMonthDay{2022}{}{}.
\newblock
{\BBOQ}\APACrefatitle {Human-Centric Artificial Intelligence Architecture for
  Industry 5.0 Applications} {Human-centric artificial intelligence
  architecture for industry 5.0 applications}.{\BBCQ}
\newblock
\APACjournalVolNumPages{arXiv preprint arXiv:2203.10794}{}{}{}.
\newblock

\newblock

\PrintBackRefs{\CurrentBib}

\bibitem [\protect \citeauthoryear {%
Ro{\v{z}}anec%
, Trajkova%
, Dam%
, Fortuna%
\BCBL {}\ \BBA {} Mladeni{\'c}%
}{%
Ro{\v{z}}anec%
, Trajkova%
\BCBL {}\ \protect \BOthers {.}}{%
{\protect \APACyear {2022}}%
}]{%
rovzanec2022streaming}
\APACinsertmetastar {%
rovzanec2022streaming}%
\begin{APACrefauthors}%
Ro{\v{z}}anec, J.M.%
, Trajkova, E.%
, Dam, P.%
, Fortuna, B.%
\BCBL {} Mladeni{\'c}, D.%
\end{APACrefauthors}%
\unskip\
\newblock
\APACrefYearMonthDay{2022}{}{}.
\newblock
{\BBOQ}\APACrefatitle {Streaming Machine Learning and Online Active Learning
  for Automated Visual Inspection.} {Streaming machine learning and online
  active learning for automated visual inspection.}{\BBCQ}
\newblock
\APACjournalVolNumPages{IFAC-PapersOnLine}{55}{2}{277--282}.
\newblock

\newblock

\PrintBackRefs{\CurrentBib}

\bibitem [\protect \citeauthoryear {%
Schmitt%
, B{\"o}nig%
, Borggr{\"a}fe%
, Beitinger%
\BCBL {}\ \BBA {} Deuse%
}{%
Schmitt%
\ \protect \BOthers {.}}{%
{\protect \APACyear {2020}}%
}]{%
schmitt2020predictive}
\APACinsertmetastar {%
schmitt2020predictive}%
\begin{APACrefauthors}%
Schmitt, J.%
, B{\"o}nig, J.%
, Borggr{\"a}fe, T.%
, Beitinger, G.%
\BCBL {} Deuse, J.%
\end{APACrefauthors}%
\unskip\
\newblock
\APACrefYearMonthDay{2020}{}{}.
\newblock
{\BBOQ}\APACrefatitle {Predictive model-based quality inspection using Machine
  Learning and Edge Cloud Computing} {Predictive model-based quality inspection
  using machine learning and edge cloud computing}.{\BBCQ}
\newblock
\APACjournalVolNumPages{Advanced engineering informatics}{45}{}{101101}.
\newblock

\newblock

\PrintBackRefs{\CurrentBib}

\bibitem [\protect \citeauthoryear {%
See%
}{%
See%
}{%
{\protect \APACyear {2012}}%
}]{%
see2012visual}
\APACinsertmetastar {%
see2012visual}%
\begin{APACrefauthors}%
See, J.E.%
\end{APACrefauthors}%
\unskip\
\newblock
\APACrefYearMonthDay{2012}{}{}.
\newblock
{\BBOQ}\APACrefatitle {Visual inspection: a review of the literature} {Visual
  inspection: a review of the literature}.{\BBCQ}
\newblock
\APACjournalVolNumPages{Sandia Report SAND2012-8590, Sandia National
  Laboratories, Albuquerque, New Mexico}{}{}{}.
\newblock

\newblock

\PrintBackRefs{\CurrentBib}

\bibitem [\protect \citeauthoryear {%
Selvi%
\ \BBA {} Nasira%
}{%
Selvi%
\ \BBA {} Nasira%
}{%
{\protect \APACyear {2017}}%
}]{%
selvi2017effective}
\APACinsertmetastar {%
selvi2017effective}%
\begin{APACrefauthors}%
Selvi, S.S.T.%
\BCBT {}\ \BBA {} Nasira, G.%
\end{APACrefauthors}%
\unskip\
\newblock
\APACrefYearMonthDay{2017}{}{}.
\newblock
{\BBOQ}\APACrefatitle {An effective automatic fabric defect detection system
  using digital image processing} {An effective automatic fabric defect
  detection system using digital image processing}.{\BBCQ}
\newblock
\APACjournalVolNumPages{J. Environ. Nanotechnol}{6}{1}{79--85}.
\newblock

\newblock

\PrintBackRefs{\CurrentBib}

\bibitem [\protect \citeauthoryear {%
Settles%
}{%
Settles%
}{%
{\protect \APACyear {2009}}%
}]{%
settles2009active}
\APACinsertmetastar {%
settles2009active}%
\begin{APACrefauthors}%
Settles, B.%
\end{APACrefauthors}%
\unskip\
\newblock
\APACrefYearMonthDay{2009}{}{}.
\newblock
{\BBOQ}\APACrefatitle {Active learning literature survey} {Active learning
  literature survey}.{\BBCQ}
\newblock

\newblock

\newblock

\PrintBackRefs{\CurrentBib}

\bibitem [\protect \citeauthoryear {%
Silva~Filho%
\ \protect \BOthers {.}}{%
Silva~Filho%
\ \protect \BOthers {.}}{%
{\protect \APACyear {2021}}%
}]{%
silva2021classifier}
\APACinsertmetastar {%
silva2021classifier}%
\begin{APACrefauthors}%
Silva~Filho, T.%
, Song, H.%
, Perello-Nieto, M.%
, Santos-Rodriguez, R.%
, Kull, M.%
\BCBL {} Flach, P.%
\end{APACrefauthors}%
\unskip\
\newblock
\APACrefYearMonthDay{2021}{}{}.
\newblock
{\BBOQ}\APACrefatitle {Classifier Calibration: How to assess and improve
  predicted class probabilities: a survey} {Classifier calibration: How to
  assess and improve predicted class probabilities: a survey}.{\BBCQ}
\newblock
\APACjournalVolNumPages{arXiv e-prints}{}{}{arXiv--2112}.
\newblock

\newblock

\PrintBackRefs{\CurrentBib}

\bibitem [\protect \citeauthoryear {%
Song%
\ \protect \BOthers {.}}{%
Song%
\ \protect \BOthers {.}}{%
{\protect \APACyear {2021}}%
}]{%
song2021classifier}
\APACinsertmetastar {%
song2021classifier}%
\begin{APACrefauthors}%
Song, H.%
, Perello-Nieto, M.%
, Santos-Rodriguez, R.%
, Kull, M.%
, Flach, P.%
\BCBL {}\ \BOthersPeriod {.}\end{APACrefauthors}%
\unskip\
\newblock
\APACrefYearMonthDay{2021}{}{}.
\newblock
{\BBOQ}\APACrefatitle {Classifier Calibration: How to assess and improve
  predicted class probabilities: a survey} {Classifier calibration: How to
  assess and improve predicted class probabilities: a survey}.{\BBCQ}
\newblock
\APACjournalVolNumPages{arXiv preprint arXiv:2112.10327}{}{}{}.
\newblock

\newblock

\PrintBackRefs{\CurrentBib}

\bibitem [\protect \citeauthoryear {%
Tsai%
\ \BBA {} Lai%
}{%
Tsai%
\ \BBA {} Lai%
}{%
{\protect \APACyear {2008}}%
}]{%
tsai2008defect}
\APACinsertmetastar {%
tsai2008defect}%
\begin{APACrefauthors}%
Tsai, D\BHBI M.%
\BCBT {}\ \BBA {} Lai, S\BHBI C.%
\end{APACrefauthors}%
\unskip\
\newblock
\APACrefYearMonthDay{2008}{}{}.
\newblock
{\BBOQ}\APACrefatitle {Defect detection in periodically patterned surfaces
  using independent component analysis} {Defect detection in periodically
  patterned surfaces using independent component analysis}.{\BBCQ}
\newblock
\APACjournalVolNumPages{Pattern Recognition}{41}{9}{2812--2832}.
\newblock

\newblock

\PrintBackRefs{\CurrentBib}

\bibitem [\protect \citeauthoryear {%
Valavanis%
\ \BBA {} Kosmopoulos%
}{%
Valavanis%
\ \BBA {} Kosmopoulos%
}{%
{\protect \APACyear {2010}}%
}]{%
valavanis2010multiclass}
\APACinsertmetastar {%
valavanis2010multiclass}%
\begin{APACrefauthors}%
Valavanis, I.%
\BCBT {}\ \BBA {} Kosmopoulos, D.%
\end{APACrefauthors}%
\unskip\
\newblock
\APACrefYearMonthDay{2010}{}{}.
\newblock
{\BBOQ}\APACrefatitle {Multiclass defect detection and classification in weld
  radiographic images using geometric and texture features} {Multiclass defect
  detection and classification in weld radiographic images using geometric and
  texture features}.{\BBCQ}
\newblock
\APACjournalVolNumPages{Expert Systems with Applications}{37}{12}{7606--7614}.
\newblock

\newblock

\PrintBackRefs{\CurrentBib}

\bibitem [\protect \citeauthoryear {%
van Garderen%
}{%
van Garderen%
}{%
{\protect \APACyear {2018}}%
}]{%
van2018active}
\APACinsertmetastar {%
van2018active}%
\begin{APACrefauthors}%
van Garderen, K.%
\end{APACrefauthors}%
\unskip\
\newblock
\APACrefYearMonthDay{2018}{}{}.
\newblock
{\BBOQ}\APACrefatitle {Active Learning for Overlay Prediction in Semi-conductor
  Manufacturing} {Active learning for overlay prediction in semi-conductor
  manufacturing}.{\BBCQ}
\newblock

\newblock

\newblock

\PrintBackRefs{\CurrentBib}

\bibitem [\protect \citeauthoryear {%
Vergara%
\ \BBA {} Est{\'e}vez%
}{%
Vergara%
\ \BBA {} Est{\'e}vez%
}{%
{\protect \APACyear {2014}}%
}]{%
vergara2014review}
\APACinsertmetastar {%
vergara2014review}%
\begin{APACrefauthors}%
Vergara, J.R.%
\BCBT {}\ \BBA {} Est{\'e}vez, P.A.%
\end{APACrefauthors}%
\unskip\
\newblock
\APACrefYearMonthDay{2014}{}{}.
\newblock
{\BBOQ}\APACrefatitle {A review of feature selection methods based on mutual
  information} {A review of feature selection methods based on mutual
  information}.{\BBCQ}
\newblock
\APACjournalVolNumPages{Neural computing and applications}{24}{1}{175--186}.
\newblock

\newblock

\PrintBackRefs{\CurrentBib}

\bibitem [\protect \citeauthoryear {%
Vergara-Villegas%
, Cruz-S{\'a}nchez%
, Jes{\'u}s Ochoa-Dom{\'\i}nguez%
, Jes{\'u}s Nandayapa-Alfaro%
\BCBL {}\ \BBA {} Flores-Abad%
}{%
Vergara-Villegas%
\ \protect \BOthers {.}}{%
{\protect \APACyear {2014}}%
}]{%
vergara2014automatic}
\APACinsertmetastar {%
vergara2014automatic}%
\begin{APACrefauthors}%
Vergara-Villegas, O.O.%
, Cruz-S{\'a}nchez, V.G.%
, Jes{\'u}s Ochoa-Dom{\'\i}nguez, H.d.%
, Jes{\'u}s Nandayapa-Alfaro, M.d.%
\BCBL {} Flores-Abad, {\'A}.%
\end{APACrefauthors}%
\unskip\
\newblock
\APACrefYearMonthDay{2014}{}{}.
\newblock
{\BBOQ}\APACrefatitle {Automatic product quality inspection using computer
  vision systems} {Automatic product quality inspection using computer vision
  systems}.{\BBCQ}
\newblock
 \APACrefbtitle {Lean Manufacturing in the Developing World} {Lean
  manufacturing in the developing world}\ (\BPGS\ 135--156).
\newblock
\APACaddressPublisher{}{Springer}.
\PrintBackRefs{\CurrentBib}

\bibitem [\protect \citeauthoryear {%
Villalba-Diez%
\ \protect \BOthers {.}}{%
Villalba-Diez%
\ \protect \BOthers {.}}{%
{\protect \APACyear {2019}}%
}]{%
villalba2019deep}
\APACinsertmetastar {%
villalba2019deep}%
\begin{APACrefauthors}%
Villalba-Diez, J.%
, Schmidt, D.%
, Gevers, R.%
, Ordieres-Mer{\'e}, J.%
, Buchwitz, M.%
\BCBL {} Wellbrock, W.%
\end{APACrefauthors}%
\unskip\
\newblock
\APACrefYearMonthDay{2019}{}{}.
\newblock
{\BBOQ}\APACrefatitle {Deep learning for industrial computer vision quality
  control in the printing industry 4.0} {Deep learning for industrial computer
  vision quality control in the printing industry 4.0}.{\BBCQ}
\newblock
\APACjournalVolNumPages{Sensors}{19}{18}{3987}.
\newblock

\newblock

\PrintBackRefs{\CurrentBib}

\bibitem [\protect \citeauthoryear {%
Villani%
}{%
Villani%
}{%
{\protect \APACyear {2009}}%
}]{%
villani2009optimal}
\APACinsertmetastar {%
villani2009optimal}%
\begin{APACrefauthors}%
Villani, C.%
\end{APACrefauthors}%
\unskip\
\newblock
\APACrefYear{2009}.
\newblock
\APACrefbtitle {Optimal transport: old and new} {Optimal transport: old and
  new}\ (\BVOL~338).
\newblock
\APACaddressPublisher{}{Springer}.
\PrintBackRefs{\CurrentBib}

\bibitem [\protect \citeauthoryear {%
Weiss%
\ \protect \BOthers {.}}{%
Weiss%
\ \protect \BOthers {.}}{%
{\protect \APACyear {2016}}%
}]{%
weiss2016continuous}
\APACinsertmetastar {%
weiss2016continuous}%
\begin{APACrefauthors}%
Weiss, S.M.%
, Dhurandhar, A.%
, Baseman, R.J.%
, White, B.F.%
, Logan, R.%
, Winslow, J.K.%
\BCBL {} Poindexter, D.%
\end{APACrefauthors}%
\unskip\
\newblock
\APACrefYearMonthDay{2016}{}{}.
\newblock
{\BBOQ}\APACrefatitle {Continuous prediction of manufacturing performance
  throughout the production lifecycle} {Continuous prediction of manufacturing
  performance throughout the production lifecycle}.{\BBCQ}
\newblock
\APACjournalVolNumPages{Journal of Intelligent Manufacturing}{27}{4}{751--763}.
\newblock

\newblock

\PrintBackRefs{\CurrentBib}

\bibitem [\protect \citeauthoryear {%
Wuest%
, Irgens%
\BCBL {}\ \BBA {} Thoben%
}{%
Wuest%
\ \protect \BOthers {.}}{%
{\protect \APACyear {2014}}%
}]{%
wuest2014approach}
\APACinsertmetastar {%
wuest2014approach}%
\begin{APACrefauthors}%
Wuest, T.%
, Irgens, C.%
\BCBL {} Thoben, K\BHBI D.%
\end{APACrefauthors}%
\unskip\
\newblock
\APACrefYearMonthDay{2014}{}{}.
\newblock
{\BBOQ}\APACrefatitle {An approach to monitoring quality in manufacturing using
  supervised machine learning on product state data} {An approach to monitoring
  quality in manufacturing using supervised machine learning on product state
  data}.{\BBCQ}
\newblock
\APACjournalVolNumPages{Journal of Intelligent
  Manufacturing}{25}{5}{1167--1180}.
\newblock

\newblock

\PrintBackRefs{\CurrentBib}

\bibitem [\protect \citeauthoryear {%
Yang%
\ \protect \BOthers {.}}{%
Yang%
\ \protect \BOthers {.}}{%
{\protect \APACyear {2020}}%
}]{%
yang2020using}
\APACinsertmetastar {%
yang2020using}%
\begin{APACrefauthors}%
Yang, J.%
, Li, S.%
, Wang, Z.%
, Dong, H.%
, Wang, J.%
\BCBL {} Tang, S.%
\end{APACrefauthors}%
\unskip\
\newblock
\APACrefYearMonthDay{2020}{}{}.
\newblock
{\BBOQ}\APACrefatitle {Using deep learning to detect defects in manufacturing:
  a comprehensive survey and current challenges} {Using deep learning to detect
  defects in manufacturing: a comprehensive survey and current
  challenges}.{\BBCQ}
\newblock
\APACjournalVolNumPages{Materials}{13}{24}{5755}.
\newblock

\newblock

\PrintBackRefs{\CurrentBib}

\bibitem [\protect \citeauthoryear {%
Yun%
, Choi%
, Jeon%
, Park%
\BCBL {}\ \BBA {} Kim%
}{%
Yun%
\ \protect \BOthers {.}}{%
{\protect \APACyear {2014}}%
}]{%
yun2014defect}
\APACinsertmetastar {%
yun2014defect}%
\begin{APACrefauthors}%
Yun, J.P.%
, Choi, D\BHBI c.%
, Jeon, Y\BHBI j.%
, Park, C.%
\BCBL {} Kim, S.W.%
\end{APACrefauthors}%
\unskip\
\newblock
\APACrefYearMonthDay{2014}{}{}.
\newblock
{\BBOQ}\APACrefatitle {Defect inspection system for steel wire rods produced by
  hot rolling process} {Defect inspection system for steel wire rods produced
  by hot rolling process}.{\BBCQ}
\newblock
\APACjournalVolNumPages{The International Journal of Advanced Manufacturing
  Technology}{70}{9}{1625--1634}.
\newblock

\newblock

\PrintBackRefs{\CurrentBib}

\bibitem [\protect \citeauthoryear {%
Zajec%
\ \protect \BOthers {.}}{%
Zajec%
\ \protect \BOthers {.}}{%
{\protect \APACyear {2021}}%
}]{%
zajec2021towards}
\APACinsertmetastar {%
zajec2021towards}%
\begin{APACrefauthors}%
Zajec, P.%
, Ro{\v{z}}anec, J.M.%
, Novalija, I.%
, Fortuna, B.%
, Mladeni{\'c}, D.%
\BCBL {} Kenda, K.%
\end{APACrefauthors}%
\unskip\
\newblock
\APACrefYearMonthDay{2021}{}{}.
\newblock
{\BBOQ}\APACrefatitle {Towards active learning based smart assistant for
  manufacturing} {Towards active learning based smart assistant for
  manufacturing}.{\BBCQ}
\newblock
 \APACrefbtitle {IFIP International Conference on Advances in Production
  Management Systems} {Ifip international conference on advances in production
  management systems}\ (\BPGS\ 295--302).
\PrintBackRefs{\CurrentBib}

\bibitem [\protect \citeauthoryear {%
Zavrtanik%
, Kristan%
\BCBL {}\ \BBA {} Sko{\v{c}}aj%
}{%
Zavrtanik%
\ \protect \BOthers {.}}{%
{\protect \APACyear {2021}}%
}]{%
zavrtanik2021draem}
\APACinsertmetastar {%
zavrtanik2021draem}%
\begin{APACrefauthors}%
Zavrtanik, V.%
, Kristan, M.%
\BCBL {} Sko{\v{c}}aj, D.%
\end{APACrefauthors}%
\unskip\
\newblock
\APACrefYearMonthDay{2021}{}{}.
\newblock
{\BBOQ}\APACrefatitle {Draem-a discriminatively trained reconstruction
  embedding for surface anomaly detection} {Draem-a discriminatively trained
  reconstruction embedding for surface anomaly detection}.{\BBCQ}
\newblock
 \APACrefbtitle {Proceedings of the IEEE/CVF International Conference on
  Computer Vision} {Proceedings of the ieee/cvf international conference on
  computer vision}\ (\BPGS\ 8330--8339).
\PrintBackRefs{\CurrentBib}

\bibitem [\protect \citeauthoryear {%
Zavrtanik%
, Kristan%
\BCBL {}\ \BBA {} Sko{\v{c}}aj%
}{%
Zavrtanik%
\ \protect \BOthers {.}}{%
{\protect \APACyear {2022}}%
}]{%
zavrtanik2022dsr}
\APACinsertmetastar {%
zavrtanik2022dsr}%
\begin{APACrefauthors}%
Zavrtanik, V.%
, Kristan, M.%
\BCBL {} Sko{\v{c}}aj, D.%
\end{APACrefauthors}%
\unskip\
\newblock
\APACrefYearMonthDay{2022}{}{}.
\newblock
{\BBOQ}\APACrefatitle {DSR--A dual subspace re-projection network for surface
  anomaly detection} {Dsr--a dual subspace re-projection network for surface
  anomaly detection}.{\BBCQ}
\newblock
 \APACrefbtitle {European Conference on Computer Vision} {European conference
  on computer vision}\ (\BPGS\ 539--554).
\PrintBackRefs{\CurrentBib}

\bibitem [\protect \citeauthoryear {%
Zeng%
\ \BBA {} Martinez%
}{%
Zeng%
\ \BBA {} Martinez%
}{%
{\protect \APACyear {2000}}%
}]{%
zeng2000distribution}
\APACinsertmetastar {%
zeng2000distribution}%
\begin{APACrefauthors}%
Zeng, X.%
\BCBT {}\ \BBA {} Martinez, T.R.%
\end{APACrefauthors}%
\unskip\
\newblock
\APACrefYearMonthDay{2000}{}{}.
\newblock
{\BBOQ}\APACrefatitle {Distribution-balanced stratified cross-validation for
  accuracy estimation} {Distribution-balanced stratified cross-validation for
  accuracy estimation}.{\BBCQ}
\newblock
\APACjournalVolNumPages{Journal of Experimental \& Theoretical Artificial
  Intelligence}{12}{1}{1--12}.
\newblock

\newblock

\PrintBackRefs{\CurrentBib}

\bibitem [\protect \citeauthoryear {%
T.~Zheng%
, Ardolino%
, Bacchetti%
\BCBL {}\ \BBA {} Perona%
}{%
T.~Zheng%
\ \protect \BOthers {.}}{%
{\protect \APACyear {2021}}%
}]{%
zheng2021applications}
\APACinsertmetastar {%
zheng2021applications}%
\begin{APACrefauthors}%
Zheng, T.%
, Ardolino, M.%
, Bacchetti, A.%
\BCBL {} Perona, M.%
\end{APACrefauthors}%
\unskip\
\newblock
\APACrefYearMonthDay{2021}{}{}.
\newblock
{\BBOQ}\APACrefatitle {The applications of Industry 4.0 technologies in
  manufacturing context: a systematic literature review} {The applications of
  industry 4.0 technologies in manufacturing context: a systematic literature
  review}.{\BBCQ}
\newblock
\APACjournalVolNumPages{International Journal of Production
  Research}{59}{6}{1922--1954}.
\newblock

\newblock

\PrintBackRefs{\CurrentBib}

\bibitem [\protect \citeauthoryear {%
Z.~Zheng%
, Zhang%
, Yu%
, Li%
\BCBL {}\ \BBA {} Zhang%
}{%
Z.~Zheng%
\ \protect \BOthers {.}}{%
{\protect \APACyear {2020}}%
}]{%
zheng2020defect}
\APACinsertmetastar {%
zheng2020defect}%
\begin{APACrefauthors}%
Zheng, Z.%
, Zhang, S.%
, Yu, B.%
, Li, Q.%
\BCBL {} Zhang, Y.%
\end{APACrefauthors}%
\unskip\
\newblock
\APACrefYearMonthDay{2020}{}{}.
\newblock
{\BBOQ}\APACrefatitle {Defect inspection in tire radiographic image using
  concise semantic segmentation} {Defect inspection in tire radiographic image
  using concise semantic segmentation}.{\BBCQ}
\newblock
\APACjournalVolNumPages{IEEE Access}{8}{}{112674--112687}.
\newblock

\newblock

\PrintBackRefs{\CurrentBib}

\end{thebibliography}


\end{document}